\documentclass[10pt,twocolumn,letterpaper]{article}

\usepackage{cvpr}              

\definecolor{cvprblue}{rgb}{0.21,0.49,0.74}
\usepackage[pagebackref,breaklinks,colorlinks,allcolors=cvprblue]{hyperref}

\usepackage{graphicx}
\usepackage{amsmath}
\usepackage{amssymb}
\usepackage{booktabs,caption}
\usepackage{multirow}
\usepackage{bm}
\usepackage{dsfont}
\usepackage[linesnumbered,ruled,vlined]{algorithm2e}
\usepackage{makecell}
\usepackage[flushleft]{threeparttable}
\usepackage{enumitem}
\usepackage[accsupp]{axessibility} 
\usepackage{algorithmic}
\usepackage[dvipsnames]{xcolor}
\definecolor{MyBlack}{HTML}{323A45}

\graphicspath{{figures/}}
\usepackage[pagebackref,breaklinks,colorlinks]{hyperref}

\usepackage[capitalize]{cleveref}
\crefname{section}{Sec.}{Secs.}
\Crefname{section}{Section}{Sections}
\Crefname{table}{Table}{Tables}
\crefname{table}{Tab.}{Tabs.}

\def\paperID{100} 
\def\confName{CVPR}
\def\confYear{2025}

\title{The Tenth NTIRE 2025 Image Denoising Challenge Report}

\author{
Lei Sun$^*$ \and
Hang Guo$^*$ \and
Bin Ren$^*$ \and
Luc Van Gool$^*$ \and
Radu Timofte$^*$ \and
Yawei Li$^*$ \and
Xiangyu Kong \and
Hyunhee Park \and
Xiaoxuan Yu \and
Suejin Han \and
Hakjae Jeon \and
Jia Li \and
Hyung-Ju Chun \and
Donghun Ryou \and
Inju Ha \and
Bohyung Han \and
Jingyu Ma \and
Zhijuan Huang \and
Huiyuan Fu \and
Hongyuan Yu \and
Boqi Zhang \and
Jiawei Shi \and
Heng Zhang \and
Huadong Ma \and
Deepak Kumar Tyagi \and
Aman Kukretti \and
Gajender Sharma \and
Sriharsha Koundinya \and
Asim Manna \and
Jun Cheng \and
Shan Tan \and
Jun Liu \and
Jiangwei Hao \and
Jianping Luo \and
Jie Lu \and
Satya Narayan Tazi \and
Arnim Gautam \and
Aditi Pawar \and
Aishwarya Joshi \and
Akshay Dudhane \and
Praful Hambadre \and
Sachin Chaudhary \and
Santosh Kumar Vipparthi \and
Subrahmanyam Murala \and
Jiachen Tu \and
Nikhil Akalwadi \and
Vijayalaxmi Ashok Aralikatti \and
Dheeraj Damodar Hegde \and
G Gyaneshwar Rao \and
Jatin Kalal \and
Chaitra Desai \and
Ramesh Ashok Tabib \and
Uma Mudenagudi \and
Zhenyuan Lin \and
Yubo Dong \and
Weikun Li \and
Anqi Li \and
Ang Gao \and
Weijun Yuan \and
Zhan Li \and
Ruting Deng \and
Yihang Chen \and
Yifan Deng \and
Zhanglu Chen \and
Boyang Yao \and
Shuling Zheng \and
Feng Zhang \and
Zhiheng Fu \and
Anas M. Ali \and
Bilel Benjdira \and
Wadii Boulila \and
JanSeny \and
Pei Zhou \and
Jianhua Hu \and
K. L. Eddie Law \and
Jaeho Lee \and
M. J. Aashik Rasool \and
Abdur Rehman \and
SMA Sharif \and
Seongwan Kim \and
Alexandru Brateanu \and
Raul Balmez \and
Ciprian Orhei \and
Cosmin Ancuti \and
Zeyu Xiao \and
Zhuoyuan Li \and
Ziqi Wang \and
Yanyan Wei \and
Fei Wang \and
Kun Li \and
Shengeng Tang \and
Yunkai Zhang \and
Weirun Zhou \and
Haoxuan Lu
}

\begin{document}

\maketitle

\let\thefootnote\relax\footnotetext{$^*$ L. Sun (lei.sun@insait.ai, INSAIT, Sofia University ``St. Kliment Ohridski''), H. Guo, B. Ren (bin.ren@unitn.it, University of Pisa \& University of Trento, Italy), L. Van Gool, R. Timofte, and Y. Li were the challenge organizers, while the other authors participated in the challenge. \\
Appendix~\ref{sec:teams} contains the authors' teams and affiliations.\\
NTIRE 2025 webpage: \url{https://cvlai.net/ntire/2025/}. \\
Code: \url{https://github.com/AHupuJR/NTIRE2025_Dn50_challenge}.
}

\begin{abstract}
This paper presents an overview of the NTIRE 2025 Image Denoising Challenge ($\sigma = 50$), highlighting the proposed methodologies and corresponding results. The primary objective is to develop a network architecture capable of achieving high-quality denoising performance, quantitatively evaluated using PSNR, without constraints on computational complexity or model size. The task assumes independent additive white Gaussian noise (AWGN) with a fixed noise level of 50. A total of 290 participants registered for the challenge, with 20 teams successfully submitting valid results, providing insights into the current state-of-the-art in image denoising.
\end{abstract}

\section{Introduction}
\label{sec:introduction}
Image denoising is a fundamental problem in low-level vision, where the objective is to reconstruct a noise-free image from its degraded counterpart.
During image acquisition and processing, various types of noise can be introduced, such as Gaussian noise, Poisson noise, and compression artifacts from formats like JPEG. The presence of these noise sources makes denoising a particularly challenging task. Given the importance of image denoising in applications such as computational photography, medical imaging, and remote sensing, continuous research efforts are necessary to develop more efficient and generalizable denoising solutions~\cite{gu2019brief,zhang2017beyond}. 


To further advance research in this area, this challenge aims to promote the development of denoising methods. A widely used benchmark for fair performance evaluation is the additive white Gaussian noise (AWGN) model, which serves as the standard setting in this competition.

As part of the New Trends in Image Restoration and Enhancement (NTIRE) 2025 workshop, we organized the Image Denoising Challenge. The objective is to restore clean images from inputs corrupted by AWGN with a noise level of $\sigma=50$. This competition seeks to foster innovative solutions, establish performance benchmarks, and explore emerging trends in the design of image denoising networks, we hope the methods in this challenge will shed light on image denoising.

This challenge is one of the NTIRE 2025~\footnote{\url{https://www.cvlai.net/ntire/2025/}} Workshop associated challenges on: ambient lighting normalization~\cite{ntire2025ambient}, reflection removal in the wild~\cite{ntire2025reflection}, shadow removal~\cite{ntire2025shadow}, event-based image deblurring~\cite{ntire2025event}, image denoising~\cite{ntire2025denoising}, XGC quality assessment~\cite{ntire2025xgc}, UGC video enhancement~\cite{ntire2025ugc}, night photography rendering~\cite{ntire2025night}, image super-resolution (x4)~\cite{ntire2025srx4}, real-world face restoration~\cite{ntire2025face}, efficient super-resolution~\cite{ntire2025esr}, HR depth estimation~\cite{ntire2025hrdepth}, efficient burst HDR and restoration~\cite{ntire2025ebhdr}, cross-domain few-shot object detection~\cite{ntire2025cross}, short-form UGC video quality assessment and enhancement~\cite{ntire2025shortugc,ntire2025shortugc_data}, text to image generation model quality assessment~\cite{ntire2025text}, day and night raindrop removal for dual-focused images~\cite{ntire2025day}, video quality assessment for video conferencing~\cite{ntire2025vqe}, low light image enhancement~\cite{ntire2025lowlight}, light field super-resolution~\cite{ntire2025lightfield}, restore any image model (RAIM) in the wild~\cite{ntire2025raim}, raw restoration and super-resolution~\cite{ntire2025raw} and raw reconstruction from RGB on smartphones~\cite{ntire2025rawrgb}.


\section{NTIRE 2025 Image Denoising Challenge}
The objectives of this challenge are threefold: (1) to stimulate advancements in image denoising research, (2) to enable a fair and comprehensive comparison of different denoising techniques, and (3) to create a collaborative environment where academic and industry professionals can exchange ideas and explore potential partnerships.

In the following sections, we provide a detailed overview of the challenge, including its dataset, evaluation criteria, challenge results, and the methodologies employed by participating teams. By establishing a standardized benchmark, this challenge aims to push the boundaries of current denoising approaches and foster innovation in the field.

\subsection{Dataset}
The widely used DIV2K~\cite{agustsson2017ntire} dataset and LSDIR~\cite{lilsdir} dataset are utilized for the challenge. 

\noindent{\textbf{DIV2K dataset}} comprises 1,000 diverse RGB images at 2K resolution, partitioned into 800 images for training, 100 images for validation, and 100 images for testing.

\noindent{\textbf{LSDIR dataset}} consists of 86,991 high-resolution, high-quality images, with 84,991 images allocated for training, 1,000 images for validation, and 1,000 images for testing.

Participants were provided with training images from both the DIV2K and LSDIR datasets. During the validation phase, the 100 images from the DIV2K validation set were made accessible to them. In the test phase, evaluation was conducted using 100 images from the DIV2K test set and an additional 100 images from the LSDIR test set. To ensure a fair assessment, the ground-truth noise-free images for the test phase remained hidden from participants throughout the challenge.

\subsection{Tracks and Competition}
The goal is to develop a network architecture that can generate high-quality denoising results, with performance evaluated based on the peak signal-to-noise ratio (PSNR) metric.

\medskip
\noindent{\textbf{Challenge phases }}
\textit{(1) Development and validation phase}: Participants were provided with 800 clean training images and 100 clean/noisy image pairs from the DIV2K dataset, along with an additional 84,991 clean images from the LSDIR dataset. During the training process, noisy images were generated by adding Gaussian noise with a noise level of $\sigma = 50$. Participants had the opportunity to upload their denoising results to the CodaLab evaluation server, where the PSNR of the denoised images was computed, offering immediate feedback on their model’s performance.
\textit{(2) Testing phase}: In the final test phase, participants were given access to 100 noisy test images from the DIV2K dataset and 100 noisy test images from the LSDIR dataset, while the corresponding clean ground-truth images remained concealed. Participants were required to submit their denoised images to the CodaLab evaluation server and send their code and factsheet to the organizers. The organizers then verified the submitted code and ran it to compute the final results, which were shared with participants at the conclusion of the challenge.

\medskip
\noindent{\textbf{Evaluation protocol}}
The primary objective of this challenge is to promote the development of accurate image denoising networks. Hence, PSNR and SSIM metrics are used for quantitative evaluation, based on the 200 test images. A code example for calculating these metrics can be found at \url{https://github.com/AHupuJR/NTIRE2025_Dn50_challenge}. Additionally, the code for the submitted solutions, along with the pre-trained weights, is also provided in this repository. Note that computational complexity and model size are not factored into the final ranking of the participants.

\section{Challenge Results}
Table~\ref{tab:rank} presents the final rankings and results of the participating teams. Detailed descriptions of each team’s implementation are provided in Sec.\ref{sec:methods_and_teams}, while team member information can be found in Appendix \ref{sec:teams}. SRC-B secured first place in terms of PSNR, achieving a 1.25 dB advantage over the second-best entry. SNUCV and BuptMM ranked second and third, respectively.

\begin{table}[t]
    \centering
    \begin{tabular}{l|c|cc}
    \hline
    Team & Rank & PSNR (primary) & SSIM  \\ \hline
    SRC-B & 1 & 31.20 & 0.8884 \\  
    SNUCV & 2 & 29.95 & 0.8676 \\  
    BuptMM & 3 & 29.89 & 0.8664 \\  
    HMiDenoise & 4 & 29.84 & 0.8653 \\  
    Pixel Purifiers & 5 & 29.83 & 0.8652 \\  
    Alwaysu & 6 & 29.80 & 0.8642 \\  
    Tcler Denoising & 7 & 29.78 & 0.8632 \\  
    cipher\_vision & 8 & 29.64 & 0.8601 \\  
    Sky-D & 9 & 29.61 & 0.8602 \\  
    KLETech-CEVI & 10 & 29.60 & 0.8602 \\  
    xd\_denoise & 11 & 29.58 & 0.8597 \\  
    JNU620 & 12 & 29.55 & 0.8590 \\  
    PSU team & 12 & 29.55 & 0.8598 \\  
    Aurora & 14 & 29.51 & 0.8605 \\  
    mpu\_ai & 15 & 29.30 & 0.8499 \\  
    OptDenoiser & 16 & 28.95 & 0.8422 \\  
    AKDT & 17 & 28.83 & 0.8374 \\  
    X-L & 18 & 26.85 & 0.7836 \\  
    Whitehairbin & 19 & 26.83 & 0.8010 \\  
    mygo & 20 & 24.92 & 0.6972 \\   \hline
    \end{tabular}
    \caption{Results of NTIRE 2025 Image Denoising Challenge. PSNR and SSIM scores are measured on the 200 test images from DIV2K test set and LSDIR test set. Team rankings are based primarily on PSNR.}
    \label{tab:rank}
\end{table}

\subsection{Participants}
\label{sec:participants}
This year, the challenge attracted 290 registered participants, with 20 teams successfully submitting valid results. Compared to the previous challenge~\cite{li2023ntire}, the SRC-B team’s solution outperformed the top-ranked method from 2023 by 1.24 dB. Notably, the results achieved by the top six teams this year surpassed those of their counterparts from the previous edition, establishing a new benchmark for image denoising.

\subsection{Main Ideas and Architectures}
\label{sec:main_ideas}
During the challenge, participants implemented a range of novel techniques to enhance image denoising performance. Below, we highlight some of the fundamental strategies adopted by the leading teams.

\begin{enumerate}
    \item \textbf{Hybrid architecture performs well.} All the models from the top-3 teams adopted a hybrid architecture that combines transformer-based and convolutional-based network. Both Global features from the transformer and local features from the convolutional network are useful for image denoising. SNUCV further adopted the model ensemble to push the limit.
    \item \textbf{Data is important.} This year's winning team, SRC-B adopted a data selection process to mitigate the influence of data imbalance, and also select high-quality images in the dataset for training instead of training on the whole DIV2K and LSDIR dataset.
    \item \textbf{The devil is in the details.} Wavelet Transform loss~\cite{korkmaz2024training} is utilized by the winning team, which is proven to help the model escape from local optima. Tricks such as a progressive learning strategy also work well. A higher percentage of overlapping of the patches during inference also leads to higher PSNR. Ensemble techniques effectively improve the performance.
    \item \textbf{New Mamba-based Design.} SNUCV, the second-ranking team, leveraged the MambaIRv2 framework to design a hybrid architecture, combining the efficient sequence modeling capabilities from Mamba with image restoration objectives.
    \item \textbf{Self-ensemble or model ensembling} is adopted to improve the performance by some of the teams.
    
\end{enumerate}

\subsection{Fairness}
\label{sec:fairness}
To uphold the fairness of the image denoising challenge, several rules were established, primarily regarding the datasets used for training. First, participants were allowed to use additional external datasets, such as Flickr2K, for training. However, training on the DIV2K validation set, including either high-resolution (HR) or low-resolution (LR) images, was strictly prohibited, as this set was designated for evaluating the generalization ability of the models. Similarly, training with the LR images from the DIV2K test set was not permitted. Lastly, employing advanced data augmentation techniques during training was considered acceptable and within the scope of fair competition.


\section{Challenge Methods and Teams}
\label{sec:methods_and_teams}



\subsection{
Samsung MX (Mobile eXperience) Business \& Samsung R\&D Institute China - Beijing (SRC-B) }

\begin{figure}
  \centering
  \includegraphics[width=0.85\linewidth]{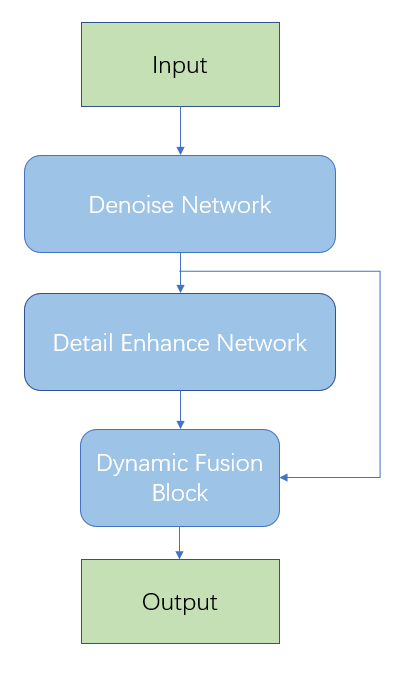}
  \caption{Framework of the hybrid network proposed by Team SRC-B.}
  \label{fig1}
\end{figure}
\subsubsection{Model Framework}
The proposed solution is shown in figure~\ref{fig1}.
In recent years, the Transformer structure has shown excellent performance in 
image denoising tasks due to its advantages in capturing global context. 

However, it is found that pure Transformer architectures are relatively weak 
in recovering local features and details. On the other hand, CNN-based methods 
excel in detail recovery but struggle to effectively capture global context 
information. Therefore, they designed a network that combines the strengths of 
the transformer network Restormer~\cite{restormer} and the convolutional 
network NAFnet~\cite{nafnet}. They first extract global 
features using the Transformer network and then enhance detail information 
using the convolutional network. 
The denoising network's structure follows Restormer, while the detail enhancement network draws inspiration from NAFNet.
Finally, they dynamically fuse the two features from 
transformer network and convolutional network 
through a set of learnable parameters to balance denoising and detail 
preservation like in , thereby improving the overall performance of image denoising.

\subsubsection{Dataset and Training Strategy}

{\bf{Dataset.}} 
Three datasets are used in total: the DIV2K dataset, the LSDIR dataset, and a self-collected custom dataset consisting of 2 million images. The specific ways in which they utilized these training sets across different training phases will be detailed in the training details section. In the final fine-tuning phase, they construct a high quality dataset consist of 1000 images from LSDIR, 1000 images from the custom dataset and all 800 images from DIV2K.  The data selection process including:
\begin{itemize}
\item Image resolution: Keep only images with a resolution greater than 900x900.
\item Image quality: Keep only images that rank in the top 30\% for all three metrics: Laplacian Var, 
BRISQUE, and NIQE. 
\item Semantic selection: To achieve semantic balance, they conducted a semantic selection based on Clip~\cite{clip} features to ensure that the dataset reflects diverse and representative content across various scene categories.
\end{itemize}

{\bf{Training.}} 
The model training consists of three stages. In the first stage, they pre-train the entire network using a custom dataset of 2 million images, with an initial learning rate of $1e^{-4}$ and a training time of approximately 360 hours. In the second stage, they fine-tune the detail enhancement network module using the DIV2K and LSDIR datasets, with an initial learning rate of $1e^{-5}$ and a training duration of about 240 hours, which enhanced the model's ability to restore details. In the third stage, they select 1,000 images from the custom dataset, 1,000 images from the LSDIR data, and 800 images from DIV2K as the training set. With an initial learning rate of $1e^{-6}$, they fine-tuned the entire network for approximately 120 hours.

The model is trained by
alternately iterating L1 loss, L2 loss, and Stationary Wavelet Transform(SWT) loss\cite{korkmaz2024training}.
They found that adding SWT loss during training helps the model escape from local optima.
They also perform progressive learning
where the network is trained on different image patch sizes
gradually enlarged from 256 to 448 and 768. As the patch
size increases, the performance can gradually improve. 
The model was trained on an A100 80G GPU.

\subsection{SNUCV}
\begin{figure*}[!h]
\includegraphics[width=1.0\linewidth]{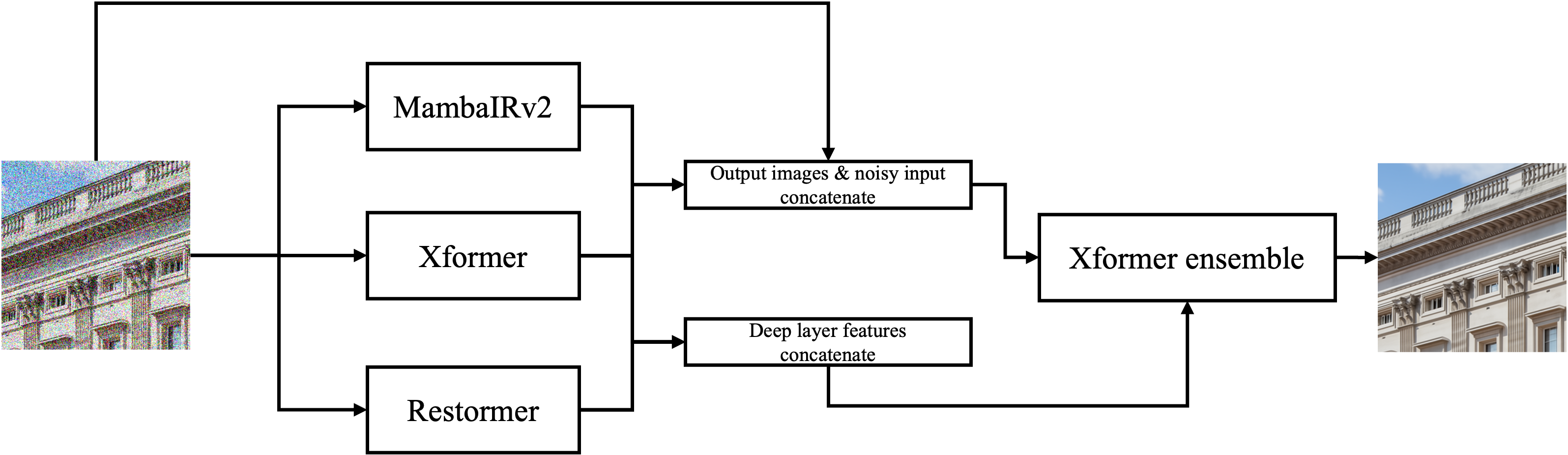}
\caption{The overview of the deep ensemble pipeline proposed by Team SNUCV.}
\label{fig:snucv_pipeline}
\end{figure*}

\textbf{Method.}
As shown in Figure~\ref{fig:snucv_pipeline}, the network architecture they utilized consists of MambaIRv2~\citep{mambairv2}, Xformer~\citep{zhang2023xformer}, and Restormer~\citep{restormer}. 
These networks were first trained on Gaussian noise with a standard deviation of 50. 
Subsequently, the outputs of these networks are concatenated with the noisy image, which is then used as input to the ensemble model. 
In addition to the output, the features from the deepest layers of these networks are also concatenated and integrated into the deepest layer features of the ensemble network. 
This approach ensures that the feature information from the previous networks is preserved and effectively transferred to the ensemble network without loss. 
The ensemble model is designed based on Xformer, accepting an input with 12 channels. 
Its deepest layer is structured to incorporate the concatenated features of the previous models. 
These concatenated features are then processed through a 1$\times$1 convolution to reduce the channel dimension back to that of the original network, thus alleviating the computational burden.
Additionally, while Xformer and Restormer reduce the feature size in their deep layer, MambaIRv2 retains its original feature size without reduction. 
To align the sizes for concatenation, the features of MambaIRv2 were downscaled by a factor of 8 before being concatenated.

\textbf{Training details.}
They first train the denoising networks, and then we incorporate the frozen denoising networks to train the ensemble model.
Both the denoising models and the ensemble model were trained exclusively using the DIV2K~\citep{agustsson2017ntire} and LSDIR~\citep{lilsdir} datasets. 
Training was performed using the AdamW~\citep{adamw} optimizer with hyperparameters \(\beta_1 = 0.9\) and \(\beta_2 = 0.999\), and a learning rate of \(3 \times 10^{-4}\). 
All models were trained for a total of 300,000 iterations.
For denoising models, Restormer and Xformer were trained using a progressive training strategy to enhance robustness and efficiency. 
Patch sizes were progressively increased as \([128, 160, 192, 256, 320, 384]\), with corresponding batch sizes of \([8, 5, 4, 2, 1, 1]\). 
In contrast, MambaIRv2 was trained with a more constrained setup due to GPU memory limitations, utilizing patch sizes of \([128, 160]\) and batch sizes of \([2, 1]\).
The ensemble model was trained with a progressive patch size schedule of \([160, 192, 256, 320, 384, 448]\) and corresponding batch sizes of \([8, 5, 4, 2, 1, 1]\).
The denoising models were trained using L1 loss, while the ensemble model was trained using a combination of L1 loss, MSE loss, and high frequency loss.

\textbf{Inference details.}
During the final inference stage to derive test results, they utilized a self-ensemble technique.
Furthermore, inference was conducted using a patch-based sliding-window approach. 
Patch sizes were set at [256, 384, 512], with corresponding overlap values of [48, 64, 96].
The resulting outputs were subsequently averaged to optimize performance.
This self-ensemble approach, while significantly increasing computational cost, substantially enhances performance.

\subsection{BuptMM}
\textbf{Description.}
In recent years, the Transformer architecture has been widely used in image denoising tasks. In order to further explore the superiority of the two representative networks, Restormer~\cite{restormer} and HAT~\cite{hat}, they propose a dual network \& post-processing denoising model that combines the advantages of the former's global attention mechanism and the latter's channel attention mechanism.

\begin{figure*}[tbp]
    \centering
    \includegraphics[width=0.95\textwidth]{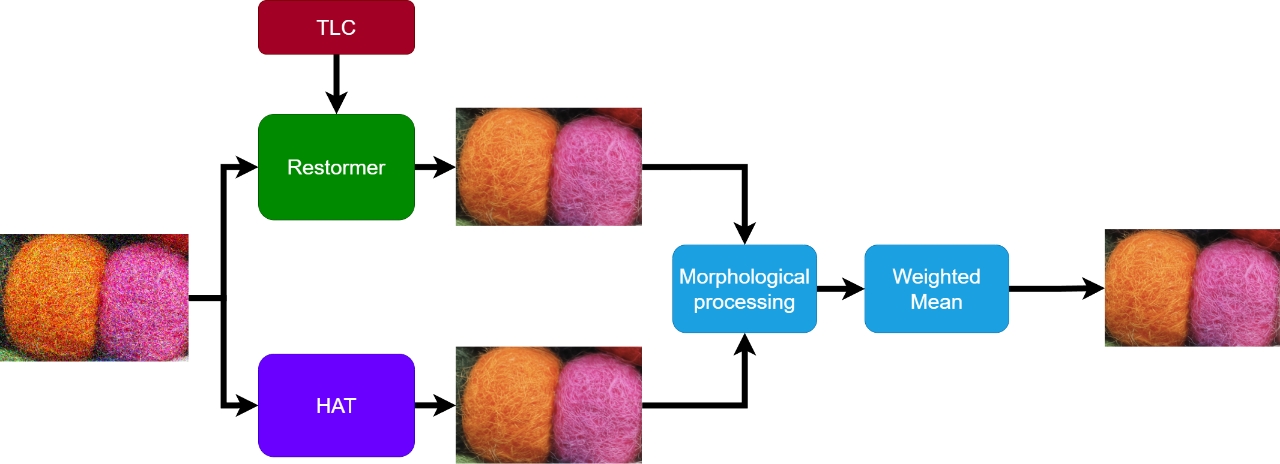}
    \caption{The model architecture of DDU proposed by Team BuptMM.}
    \label{fig:buptmm}
\end{figure*}

As shown in Fig.~\ref{fig:buptmm}, our network is divided into two stages. In the first stage, they use DIV2K~\cite{agustsson2017ntire}  and LSDIR~\citep{lilsdir} training sets to train Restormer~\cite{restormer} and HAT~\cite{hat} respectively, and then enhance the ability of Restormer~\cite{restormer} through TLC~\cite{lin2024improving} technology during its reasoning stage. In the second stage, they first use the Canny operator to perform edge detection on the images processed by the two models. They take an OR operation on the two edge images, and then XOR the result with the edge of HAT to obtain the edge difference between the two images. For this part of the edge difference, they use the result obtained by HAT~\cite{hat} as the standard for preservation. Finally, they take the average of the other pixels of HAT~\cite{hat} and Restormer~\cite{restormer} to obtain the final result.

They used the DIV2K~\cite{agustsson2017ntire} and LSDIR~\cite{lilsdir} datasets to train both the Restormer~\cite{restormer} and HAT~\cite{hat} simultaneously. They employed a progressive training strategy for the Restormer~\cite{restormer}with a total of 292000 iterations, where the image block size increased from 128 to 384 with a step size of 64. They also used progressive training strategy for the HAT~\cite{hat}, where the image block size increased from 64 to 224. They did not use any other datasets besides the datasets mentioned above during the process. During the training phase, they spent one day separately training the Reformer~\cite{restormer} and HAT~\cite{hat}, they trained two models using 8 NVIDIA H100 GPUs. They conducted the inference process on the H20 test set, with a memory usage of 15G. The average inference time for a single image from the 200 test sets was 4.4 seconds, while the average time for morphological post-processing was within 1 second.
\subsection{HMiDenoise}
\begin{figure}[t]
    \centering
    \includegraphics[width=0.4\textwidth]{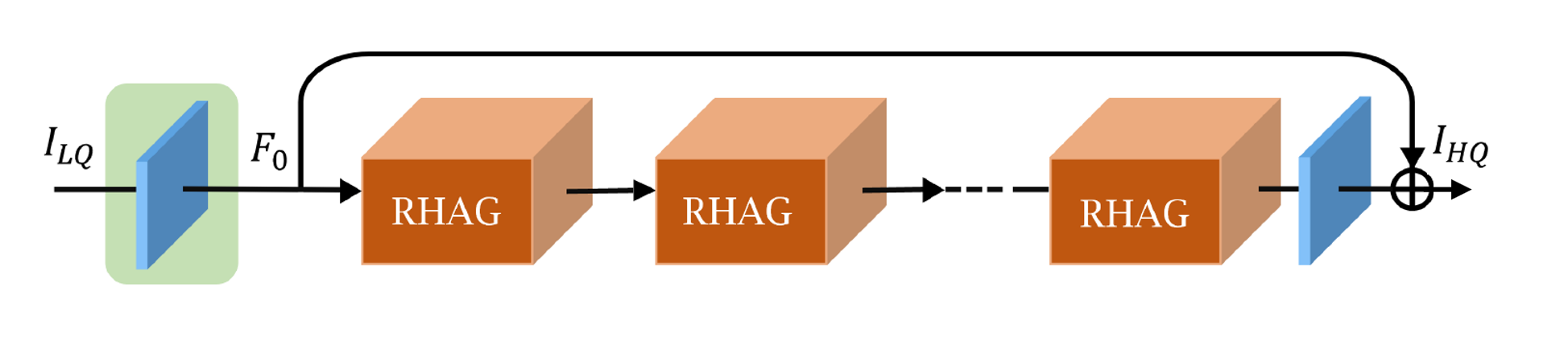}
    \caption{Model architecture of DB-HAT proposed by Team HMiDenoise.}
    \label{fig:denoisingHAT_network}
\end{figure}

The network is inspired by the HAT~\cite{hat} model architecture, and the architecture is optimized for the task specifically. The optimized denoising network structure(D-HAT) is shown in Fig~\ref{fig:denoisingHAT_network}. 
 
The dataset utilized for training comprises DIV2K and LSDIR. To accelerate training and achieve good performance, they initially train on a small scale (64x64) with batch size 16, then on a medium scale (128x128) with batch size 1, and finally optimize on a larger scale (224x224) with batch size 1. As the patch size increases, the performance can gradually improve. The learning rate is initialized at  4 × $10^{-4}$ and decays according to the cosine annealing strategy during the training. The network undergoes training for a total of 2×$10^{5}$ iterations, with the L2 loss function being minimized through the utilization of the Adam optimizer. Subsequently, fine-tuning is executed using the L2 loss and SSIM loss functions, with an initial learning rate of 5 × $10^{-5}$ for 2×$10^{5}$ iterations. They repeated the aforementioned fine-tune settings two times after loading the trained weights. All experiments are conducted with the PyTorch 2.0 framework on 8 H100 GPUs.

\subsection{Pixel Purifiers}

\begin{figure*}[htbp]    \centering    \includegraphics[width=0.9\textwidth]{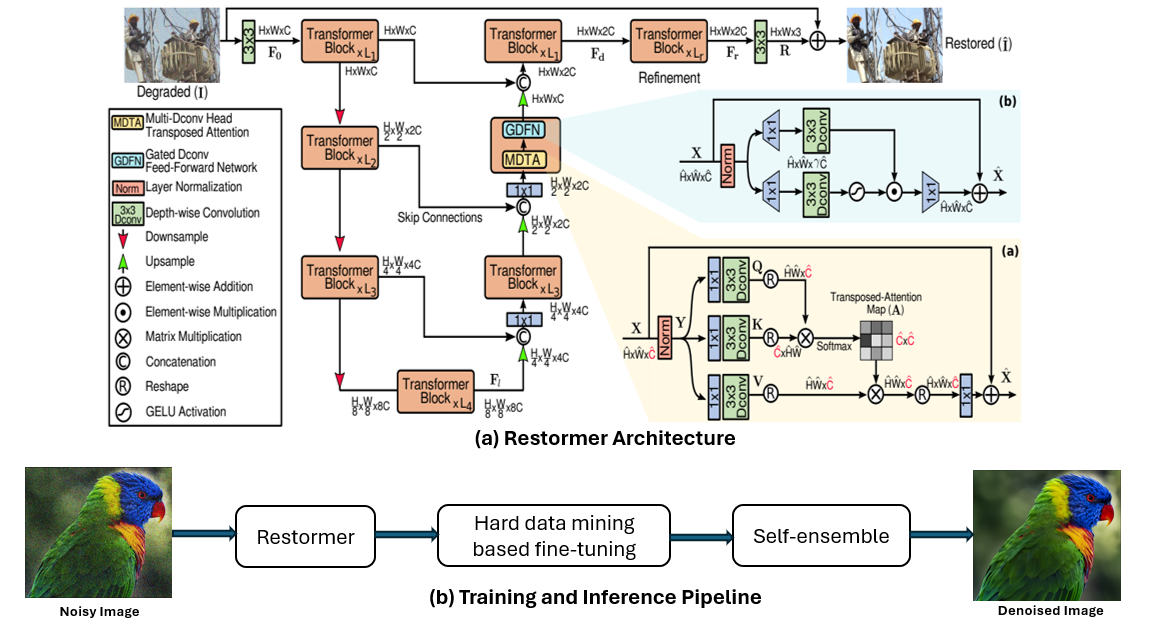}    \caption{Block Diagram for Image Denoising using Restormer architecture along with Hard data mining and Ensemble Techniques (Team Pixel Purifiers).}    \label{fig:Pixel Purifiers}\end{figure*}

\textbf{Architecture.} Restormer architecture \cite{restormer}, as shown in Fig.~\ref{fig:Pixel Purifiers}(a), is an efficient transformer and it uses the multi-Dconv head transposed attention block (MDTA) for channel attention and the gated Dconv feedforward network (GDFN) for the feedforward network. MDTA block applies self-attention across channels rather than the spatial dimension to compute cross-covariance across channels to generate an attention map encoding the global context implicitly. Additionally, depth-wise convolutions are used to emphasize on the local context before computing feature covariance to produce the global attention map. GDFN block introduces a novel gating mechanism and depth-wise convolutions to encode information from spatially neighboring pixel positions, useful for learning local image structure for effective restoration.

\textbf{Training Techniques.} They have conducted extensive experiments to evaluate the effectiveness of our approach (as shown in Fig.~\ref{fig:Pixel Purifiers}(b)). The network is trained using the DIV2K and LSDIR datasets only with L1 loss function. To enhance generalization and mitigate overfitting, they apply randomized data augmentation during training, including horizontal flipping, vertical flipping, and rotations of $90^{\circ}$, $180^{\circ}$, and $270^{\circ}$. A fixed patch size of \(256\times256\) is maintained for both training and inference to preserve global context. For optimization, they used the AdamW optimizer in conjunction with the CosineAnnealingRestartCyclicLR scheduler, with an initial learning rate $1 \times 10^{-4}$. Training is done using 8 NVIDIA Tesla V100 GPUs. Additionally, they leveraged Hard Dataset Mining for model fine-tuning, specifically targeting training patches where the loss exceeded a predefined threshold. This technique, discussed in detail in the following section, further enhanced the performance of our baseline model. 

\textbf{Hard Dataset Mining.} To further enhance PSNR, they employed a hard dataset mining technique inspired by \cite{becker2023make} for fine-tuning. Specifically, training patches with loss value exceeding a predefined threshold is selected for transfer learning on our base trained model. To preserve the model’s generalization while refining its performance on challenging samples, they applied a learning rate that was 100 times smaller than the initial training rate.

\textbf{DIV2K and LSDIR Datasets Ratio. } As the model is to be trained and tested on two datasets (DIV2K and LSDIR), they first analysed their characteristics. DIV2K is relatively small and generalised with 800 training images while LSDIR is significantly large dataset with 84k+ training images, primarily consisting of high texture images. Considering the dataset charatectertics and our dataset ratio experiments, they found that DIV2K to LSDIR ratio of 12:88 during training helps to improve overall PSNR and generalise the model better for both validation and test datasets.

\textbf{Overlapping Percentage During Inference. } Using a small overlap of 5\% during inference with a patch size of \(256\times256\) (same as the training patch size to preserve global context) resulted in improved inference speed. However, despite applying boundary pixel averaging, minor stitching artifacts is observed, leading to a decline in PSNR performance. To mitigate these artifacts, they increased the overlap to 20\% with original \(256\times256\) patch size, which resulted in PSNR improvement.

\textbf{Ensemble Technique at Inference. } Ensemble techniques played a crucial role by effectively boosting performance. They used the Self Ensemble Strategy, specifically test-time augmentation ensemble \cite{lim2017enhanced} where multiple flips and rotations of images were used before model inference. The model outputs are averaged to generate the final output image.

\subsection{Alwaysu}
\noindent \textbf{Method}: Our objective is to achieve efficient Gaussian denoising based on pre-trained denoisers. Our core idea, termed \textit{Bias-Tuning}, initially proposed in transfer learning \cite{cai2020tinytl}, is freezing pre-trained denoisers and 
only fine-tuning \textit{existing or newly added bias parameters} during adaptation, thus maintaining the knowledge of pre-trained models and reducing tuning cost.

They choose the Restormer \cite{restormer} model trained to remove the same $i.i.d.$ Gaussian noise ($\sigma=50$) \textit{without intensity clipping} as our baseline. As this pre-trained Restormer did not clip noisy images' intensities into the normal range, i.e., [0, 255], it performs poorly in \textit{clipped} noisy images, resulting in low PSNR/SSIM (27.47/0.79 on DIV2K validation) and clear artifacts. After embedding learnable bias parameters into this freezing Restormer (except LayerNorm modules) and fine-tuning the model, satisfactory denoising results can be obtained, and the resultant PSNR increases by over 3dB (evaluated on DIV2K validation set). They found that various pre-trained Gaussian denoisers from \cite{restormer}, including three noise-specific models and one noise-blind model, resulted in similar denoising performance on clipped noisy images after Bias-Tuning. 

During the inference, they further enhance the denoiser via self-ensemble \cite{lim2017enhanced} and \textit{patch stitching}. When dealing with high-resolution (HR) noisy images, they process them via overlapping patches with the same patch size as the training phase. They stitch these overlapping denoised patches via linear blending, as introduced in image stitching \cite{brown2007automatic}. 

\noindent \textbf{Training details}: They fine-tune this \textit{bias-version} Restormer using the PSNR loss function and AdamW optimizer combined with batch size $2$, patch size $256\times 256$, learning rate $3e^{-4}$ (cosine annealed to $1e^{-6}$), $200k$ iterations and geometric augmentation. The training dataset consists of 800 images from DIV2K training set and 1,000 images from LSDIR training set. They also note that the pre-trained Restormer utilized a combined set of 800 images from DIV2K, 2,650 images of Flickr2K, 400 BSD500 images and 4,744 images from WED.

\noindent \textbf{Inference details}: 
The patch size and overlapping size during patch stitching are $256\times256$ and $16$, respectively.

\noindent \textbf{Complexity}: Total number of parameters: 26.25M; Total number of learnable bias parameters: 0.014M; FLOPs: 140.99G (evaluated on image with shape $256\times256\times3$).
\subsection{Tcler\_Denosing}
Building upon the work of Potlapalli et al. \cite{potlapalli2023promptir}, they propose a novel transformer-based architecture for image restoration, termed PromptIR-Dn50. This architecture adopts a U-shaped encoder-decoder network structure, incorporating progressive downsampling and upsampling operations. Specifically tailored for denoising tasks under additive white Gaussian noise (AWGN) with a noise level of sigma=50, PromptIR-Dn50 leverages the strengths of the PromptGenBlock with targeted modifications. In this framework, the PromptGenBlock is adapted by explicitly incorporating sigma=50 as an input parameter, ensuring the model is optimized for the specific noise level and achieves superior performance in denoising tasks.

Inspired by the advancements in MambaIRv2 \cite{mambairv2}, they further introduce a specialized variant, MambaIRv2-Dn50, designed for image restoration tasks. This architecture also adopts a U-shaped encoder-decoder structure but integrates two key innovations: the Attentive State-space Equation (ASE) and Semantic Guided Neighboring (SGN) modules. These components address the causal scanning limitations inherent in traditional Mamba frameworks while maintaining linear computational complexity. Unlike prior approaches that rely on multi-directional scanning, MambaIRv2-Dn50 achieves non-causal global perception through single-sequence processing, making it highly efficient and well-suited for vision tasks.

To further enhance the performance of image restoration, they propose a fusion strategy that combines the strengths of PromptIR-Dn50 and MambaIRv2-Dn50. By integrating the outputs of these two architectures, the fused model leverages the noise-specific optimization of PromptIR-Dn50 and the global perception capabilities of MambaIRv2-Dn50. This hybrid approach ensures robust and high-quality restoration results, effectively addressing the challenges posed by sigma=50 AWGN noise.

The architecture follows a progressive training strategy as in Restormer \cite{restormer}, where input resolutions gradually increase from 64×64 to 112×112. This progressive learning scheme enhances feature adaptation across scales without compromising training stability.

For optimization, they employ the Adam optimizer with an initial learning rate of 1e-4, combined with a CosineAnnealingRestartCyclicLR schedule to adjust the learning rate dynamically during training. The model is trained using a combination of Charbonnier loss and Gradient-weighted L1 loss, which effectively balances pixel-wise accuracy and edge preservation. The weights for those two losses are 0.8 and 0.2, respectively. They use the DIV2K \cite{agustsson2017ntire} and LSDIR \cite{lilsdir}  datasets exclusively during the training phase, where horizontally and vertically flipping, rotation, USM sharpen \cite{wang2021real} are used to augment the input images of our model.

During the testing phase, the input size is fixed at 112×112, and self-ensemble techniques \cite{timofte2016seven} are applied to further enhance the model's performance. This approach ensures robust denoising results and improved generalization to unseen data.

In summary, MambaIRv2-Dn50 introduces a tailored state-space model-based architecture for denoising tasks, leveraging progressive learning, advanced loss functions, and self-ensemble techniques to achieve state-of-the-art performance on sigma=50 AWGN noise.
\subsection{cipher\_vision}
\begin{figure}[t] 
\centering
\includegraphics[width=\linewidth]{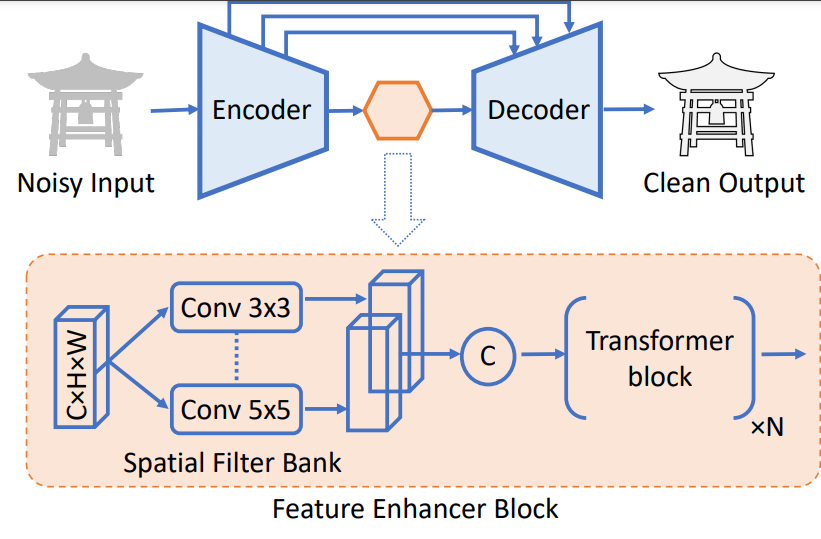} 
\caption{Proposed Pureformer encoder-decoder architecture for image denoising proposed by Team cipher vision. The input noisy image is processed through a multi-level encoder, a feature enhancer block, and a multi-level decoder. Each encoder and decoder level employs \textit{xN} transformer blocks \cite{zhang2023practical}, consisting of Multi-Dconv Head Transposed Attention (MDTA) and Gated-Dconv Feed-Forward Network (GDFN) blocks. The feature enhancer block, placed in the latent space, expands the receptive field using a spatial filter bank. The multi-scale features are then concatenated and refined through \textit{xN} transformer blocks to enhance feature correlation and merge multi-scale information effectively.}
\label{fig:Pureformer}
\end{figure}

As shown in Figure~\ref{fig:Pureformer}, they employ a Transformer-based encoder-decoder architecture featuring a four-level encoder-decoder structure designed to restore images degraded by Gaussian noise ($\sigma = 50$). This architecture is optimized to capture both local and global features, significantly enhancing the quality of input images. The hierarchical structure of the model includes four levels, containing [4, 6, 6, 8] Transformer blocks respectively. Each Transformer block includes Multi-Dconv Head Transposed Attention (MDTA) followed by a Gated-Dconv feed-forward network (GDFN), enabling the model to capture long-range feature dependencies effectively. Additionally, skip connections are utilized to link the encoder and decoder, preserving spatial details and ensuring efficient feature reuse throughout the network. The feature enhancer block in the latent space processes latent features through the filter bank, and extracted multi-scale features are concatenated and passed through the transformer blocks as shown in Figure~\ref{fig:Pureformer}.

\textbf{Training Details}
Our training strategy uses the datasets DIV2K (1000) and LSDIR (86,991). They leverage small patch-based training and data augmentation techniques to optimize the Pureformer. The training process uses the AdamW optimizer ($\beta_1 = 0.9$, $\beta_2 = 0.999$) with a learning schedule that includes a linear warmup for 15 epochs followed by cosine annealing. The batch size is set to 4, consisting of $4 \times 3 \times 128 \times 128$ patches, and training is conducted on 2×A100 GPUs. Data augmentation techniques such as random cropping, flips, 90° rotations, and mixup are applied. They use L1 Loss to optimize the parameters.

\textbf{Testing Strategy}
For inference, they use the datasets DIV2K (100) and LSDIR (100). Testing is performed using $512 \times 512$ patches. To enhance robustness, they employ self-ensemble testing with rotational transformations. The input image is rotated by 0°, 90°, 180°, and 270°, processed through the trained model, and rotated back to its original orientation. The final prediction is obtained by averaging the outputs of all four rotations.

\subsection{A Two-Stage Denoising Framework with Generalized Denoising Score Matching Pretraining and Supervised Fine-tuning (Sky-D)}
\paragraph{Problem Formulation}
In natural image denoising, we aim to recover a clean image $\mathbf{X}_0 \in \mathbb{R}^d$ from its noisy observation $\mathbf{X}_{t_{\text{data}}} \in \mathbb{R}^d$. The noisy observation can be modeled as:
\begin{equation}
    \mathbf{X}_{t_{\text{data}}} = \mathbf{X}_0 + \sigma_{t_{\text{data}}} \mathbf{N},
    \label{eq:noisy_model}
\end{equation}
where $\sigma_{t_{\text{data}}} > 0$ denotes the noise standard deviation at level $t_{\text{data}}$, and $\mathbf{N} \sim \mathcal{N}(\mathbf{0}, \mathbf{I}_d)$ represents the noise component.

Our approach consists of two stages: (1) self-supervised pretraining using Generalized Denoising Score Matching (GDSM) and (2) supervised fine-tuning. This two-stage approach enables us to leverage both noisy data and clean labels effectively.

\subsubsection{Self-Supervised Pretraining with Generalized Denoising Score Matching}

For the pretraining stage, we adopt the Generalized Denoising Score Matching (GDSM) framework introduced in Corruption2Self (C2S) \cite{tu2025scorebased}. This approach enables effective learning directly from noisy observations without requiring clean labels.

\paragraph{Forward Corruption Process}

Following \cite{tu2025scorebased}, we define a forward corruption process that systematically adds additional Gaussian noise to $\mathbf{X}_{t_{\text{data}}}$:

\begin{equation}
\begin{aligned}
    \mathbf{X}_t &= \mathbf{X}_{t_{\text{data}}} + \sqrt{\sigma_t^2 - \sigma_{t_{\text{data}}}^2} \, \mathbf{Z}, \\
    \mathbf{Z} &\sim \mathcal{N}(\mathbf{0}, \mathbf{I}_d), \quad t > t_{\text{data}},
\end{aligned}
    \label{eq:forward_process}
\end{equation}

where $\sigma_t$ is a monotonically increasing noise schedule function for $t \in (t_{\text{data}}, T]$, with $T$ being the maximum noise level.

\paragraph{Generalized Denoising Score Matching Loss}

The GDSM loss function \cite{tu2025scorebased} is formulated as:

\begin{equation}
\begin{aligned}
    J(\theta) = \mathbb{E}_{\mathbf{X}_{t_{\text{data}}}, t, \mathbf{X}_t} \Big[ \big\| \gamma(t, \sigma_{t_{\text{target}}}) \, \mathbf{h}_\theta(\mathbf{X}_t, t) \\
    + \delta(t, \sigma_{t_{\text{target}}}) \, \mathbf{X}_t - \mathbf{X}_{t_{\text{data}}} \big\|^2 \Big],
\end{aligned}
    \label{eq:general_loss}
\end{equation}

where $t$ is sampled uniformly from $(t_{\text{data}}, T]$ and the coefficients are defined by:

\begin{equation}
\begin{aligned}
\gamma(t, \sigma_{t_{\text{target}}}) &:= \frac{\sigma_t^2 - \sigma_{t_{\text{data}}}^2}{\sigma_t^2 - \sigma_{t_{\text{target}}}^2} \\
\delta(t, \sigma_{t_{\text{target}}}) &:= \frac{\sigma_{t_{\text{data}}}^2 - \sigma_{t_{\text{target}}}^2}{\sigma_t^2 - \sigma_{t_{\text{target}}}^2}.
\end{aligned}
\end{equation}

The parameter $\sigma_{t_{\text{target}}}$ controls the target noise level, with $\sigma_{t_{\text{target}}} = 0$ representing maximum denoising (complete noise removal).

\paragraph{Reparameterization for Improved Training Stability}

To enhance training stability and improve convergence, we employ the reparameterization strategy proposed in \cite{tu2025scorebased}. Let $\tau \in (0, T']$ be a new variable defined by:

\begin{equation}
\begin{aligned}
\sigma_\tau^2 &= \sigma_t^2 - \sigma_{t_{\text{data}}}^2, \\
T' &= \sqrt{\sigma_T^2 - \sigma_{t_{\text{data}}}^2}.
\end{aligned}
\end{equation}

The original $t$ can be recovered via:
\begin{equation}
t = \sigma_t^{-1}\left( \sqrt{ \sigma_\tau^2 + \sigma_{t_{\text{data}}}^2 } \right).
\end{equation}

Under this reparameterization, the loss function becomes:
\begin{equation}
\begin{aligned}
    J'(\theta) = \mathbb{E}_{\mathbf{X}_{t_{\text{data}}}, \tau, \mathbf{X}_t} \Big[ \big\| \gamma'(\tau, \sigma_{t_{\text{target}}}) \, \mathbf{h}_\theta(\mathbf{X}_t, t) \\
    + \delta'(\tau, \sigma_{t_{\text{target}}}) \, \mathbf{X}_t - \mathbf{X}_{t_{\text{data}}} \big\|^2 \Big],
\end{aligned}
    \label{eq:reparam_loss}
\end{equation}

where the coefficients are:
\begin{equation}
\begin{aligned}
\gamma'(\tau, \sigma_{t_{\text{target}}}) &= \frac{\sigma_{\tau}^2}{\sigma_{\tau}^2 + \sigma_{t_{\text{data}}}^2 - \sigma_{t_{\text{target}}}^2}, \\
\delta'(\tau, \sigma_{t_{\text{target}}}) &= \frac{\sigma_{t_{\text{data}}}^2 - \sigma_{t_{\text{target}}}^2}{\sigma_{\tau}^2 + \sigma_{t_{\text{data}}}^2 - \sigma_{t_{\text{target}}}^2}.
\end{aligned}
\end{equation}

This reparameterization ensures uniform sampling over $\tau$ and consistent coverage of the noise level range during training, leading to smoother and faster convergence.

\subsubsection{Supervised Fine-tuning}

After pretraining with GDSM, we propose to fine-tune the model with a supervised approach. Unlike traditional methods that train from scratch using clean labels, our approach leverages the knowledge gained during pretraining to enhance performance.

\paragraph{Supervised Fine-tuning Loss}

Given paired training data $\{(\mathbf{X}_{t_{\text{data}}}^i, \mathbf{Y}^i)\}_{i=1}^N$, where $\mathbf{X}_{t_{\text{data}}}^i$ is the noisy observation and $\mathbf{Y}^i$ is the corresponding clean target, we formulate the supervised fine-tuning loss as:

\begin{equation}
    \mathcal{L}_{\text{sup}}(\theta) = \frac{1}{N} \sum_{i=1}^N \left\| \mathbf{h}_\theta(\mathbf{X}_{t_{\text{data}}}^i, t_{\text{data}}) - \mathbf{Y}^i \right\|^2.
    \label{eq:supervised_loss}
\end{equation}

This formulation directly optimizes the network to map noisy observations to clean targets. By initializing $\theta$ with the pretrained weights from the GDSM stage, we enable more effective and stable fine-tuning.

\subsubsection{Time-Conditioned Diffusion Model Architecture}

Our approach employs the same time-conditioned diffusion model architecture used in \cite{tu2025scorebased}, which is based on the U-Net architecture enhanced with time conditioning and the Noise Variance Conditioned Multi-Head Self-Attention (NVC-MSA) module. The model's denoising function $\mathbf{h}_\theta : \mathbb{R}^d \times \mathbb{R} \to \mathbb{R}^d$ maps a noisy input $\mathbf{X}_t$ and noise level $t$ to an estimate of the clean image $\mathbf{X}_0$.

The time conditioning is implemented through an embedding layer that transforms the noise level $t$ into a high-dimensional feature vector, which is then integrated into the convolutional layers via adaptive instance normalization. This enables the model to dynamically adjust its denoising behavior based on the noise level of the input.

The NVC-MSA module extends standard self-attention by conditioning the attention mechanism on the noise variance, allowing the model to adapt its attention patterns based on the noise characteristics of the input. This adaptation enhances the model's ability to denoise effectively across different noise levels and patterns.

\subsubsection{Training Procedure}
\begin{algorithm}[t]
\caption{Two-Stage Training Procedure for GDSM Pretraining and Supervised Fine-tuning}
\label{alg:two_stage_training}
\begin{algorithmic}[1]
\small
\REQUIRE Training data from DIV2K and LSDIR, max noise level $T$, learning rates $\alpha_1$, $\alpha_2$
\ENSURE Trained denoising model $\mathbf{h}_\theta$

\STATE \textbf{// Phase 1: Self-supervised Pretraining with GDSM}
\STATE Initialize network parameters $\theta$ randomly
\REPEAT
    \STATE Sample minibatch $\{\mathbf{X}_{t_{\text{data}}}^i\}_{i=1}^m$ from DIV2K and LSDIR training sets
    \STATE Sample noise level $\tau \sim \mathcal{U}(0, T']$
    \STATE Sample Gaussian noise $\mathbf{Z} \sim \mathcal{N}(\mathbf{0}, \mathbf{I}_d)$
    \STATE Compute $t = \sigma_t^{-1}\left( \sqrt{\sigma_\tau^2 + \sigma_{t_{\text{data}}}^2} \right)$
    \STATE Generate corrupted samples: $\mathbf{X}_t = \mathbf{X}_{t_{\text{data}}} + \sigma_{\tau} \mathbf{Z}$
    \STATE Compute coefficients $\gamma'(\tau, \sigma_{t_{\text{target}}})$ and $\delta'(\tau, \sigma_{t_{\text{target}}})$
    \STATE Compute GDSM loss $J'(\theta)$ according to Eq.~\eqref{eq:reparam_loss}
    \STATE Update parameters: $\theta \leftarrow \theta - \alpha_1 \nabla_\theta J'(\theta)$
\UNTIL{convergence or maximum iterations reached}

\STATE \textbf{// Phase 2: Supervised Fine-tuning}
\STATE Initialize network parameters $\theta$ with pretrained weights from Phase 1
\REPEAT
    \STATE Sample paired minibatch $\{(\mathbf{X}_{t_{\text{data}}}^i, \mathbf{Y}^i)\}_{i=1}^m$ from DIV2K and LSDIR training sets
    \STATE Compute supervised loss: $\mathcal{L}_{\text{sup}}(\theta) = \frac{1}{m} \sum_{i=1}^m \|\mathbf{h}_\theta(\mathbf{X}_{t_{\text{data}}}^i, t_{\text{data}}) - \mathbf{Y}^i\|^2$
    \STATE Update parameters: $\theta \leftarrow \theta - \alpha_2 \nabla_\theta \mathcal{L}_{\text{sup}}(\theta)$ \COMMENT{$\alpha_2 < \alpha_1$ for stable fine-tuning}
\UNTIL{convergence or maximum iterations reached}

\RETURN Trained model $\mathbf{h}_\theta$
\end{algorithmic}
\end{algorithm}

As outlined in Algorithm~\ref{alg:two_stage_training}, our approach combines self-supervised pretraining with supervised fine-tuning to leverage the strengths of both paradigms. The GDSM pretraining phase enables the model to learn robust representations across diverse noise levels without clean labels, establishing a strong initialization for subsequent supervised learning. This knowledge transfer accelerates convergence during fine-tuning and enhances generalization to noise distributions not explicitly covered in the supervised data. The time-conditioned architecture further facilitates this adaptability by dynamically adjusting denoising behavior based on input noise characteristics. To our knowledge, this represents the first application of GDSM as a pretraining strategy for natural image denoising, offering a principled approach to combining self-supervised and supervised learning objectives for this task.

\subsubsection{Implementation Details}

We implement our two-stage training procedure with a progressive learning strategy similar to that proposed in \cite{restormer}, gradually increasing image patch sizes to capture multi-scale features while maintaining computational efficiency. As detailed in Algorithm~\ref{alg:two_stage_training}, each stage consists of both self-supervised pretraining and supervised fine-tuning phases.

For the GDSM pretraining, we set the maximum corruption level $T=10$, which provides sufficient noise coverage while maintaining training stability. To determine the data noise level $t_{\text{data}}$, we incorporate standard noise estimation techniques from the \texttt{skimage} package \cite{van2014scikit}. While we could explicitly set $t_{\text{data}}$ to correspond to specific noise levels (e.g., 50/255), we found that automated estimation suffices for good performance. In future work, more tailored approaches for specific noise level denoising could be implemented.

For optimization, we employ the AdamW optimizer with gradient clipping to stabilize training, coupled with a cosine annealing learning rate scheduler. Our progressive training schedule (see Table~\ref{tab:progressive_training}) gradually increases patch sizes while adjusting batch sizes and learning rates accordingly. We initialize each stage with weights from the previous stage, setting a maximum of 20 epochs per stage with early stopping based on validation performance. Due to computational time constraints, we note that the network training for the final stage of progressive learning had not yet fully converged when reporting our results.

\begin{table}[t]
\centering
\small
\caption{Progressive Training Schedule}
\label{tab:progressive_training}
\begin{tabular}{@{}lccc@{}}
\toprule
\textbf{Stage} & \textbf{Patch Size} & \textbf{Batch} & \textbf{Learning Rate} \\
\midrule
1 & $256^2$ & 48 & $1 \times 10^{-3}$ \\
2 & $384^2$ & 24 & $3 \times 10^{-4}$ \\
3 & $512^2$ & 12 & $1 \times 10^{-4}$ \\
4 & Mixed* & 4 & $5 \times 10^{-5}$ \\
\bottomrule
\end{tabular}
\\
\footnotesize{*Randomly selected from $\{512^2, 768^2, 896^2\}$ per batch}
\end{table}

This progressive approach allows the model to initially learn basic denoising patterns on smaller patches where more diverse samples can be processed in each batch, then gradually adapt to larger contextual information in later stages. We train our models using the DIV2K \cite{agustsson2017ntire} and LSDIR \cite{lilsdir} training datasets, while validation is performed on their respective validation sets, which remain completely separate from training.

Throughout the entire training process, we maintain the same time-conditioned model architecture, leveraging its ability to handle varying noise levels both during self-supervised pretraining and supervised fine-tuning. The self-supervised pretraining with GDSM establishes robust initialization across diverse noise conditions, while the supervised fine-tuning further refines the model's performance on specific noise distributions of interest.

\subsubsection{Inference Process}

During standard inference, given a noisy observation $\mathbf{X}_{t_{\text{data}}}$, we obtain the denoised output directly from our trained model:

\begin{equation}
\hat{\mathbf{X}} = \mathbf{h}_{\theta^*}(\mathbf{X}_{t_{\text{data}}}, t_{\text{data}}),
\end{equation}

However, to maximize denoising performance for high-resolution images without requiring additional model training, we incorporate two advanced techniques: geometric self-ensemble and adaptive patch-based processing.

\paragraph{Geometric Self-Ensemble}

Following \cite{lim2017enhanced}, we implement geometric self-ensemble to enhance denoising quality by leveraging the model's equivariance properties. This technique applies a set of geometric transformations (rotations and flips) to the input image, processes each transformed version independently, and then averages the aligned outputs. The approach can be concisely formulated as:

\begin{equation}
\hat{\mathbf{X}}_{\text{GSE}} = \frac{1}{K}\sum_{i=1}^{K} T_i^{-1}\left(\mathbf{h}_{\theta^*}\left(T_i(\mathbf{X}_{t_{\text{data}}}), t_{\text{data}}\right)\right),
\end{equation}

where $\{T_i\}_{i=1}^K$ represents a set of $K=8$ geometric transformations (identity, horizontal flip, vertical flip, 90°, 180°, and 270° rotations, plus combinations), and $T_i^{-1}$ denotes the corresponding inverse transformation. This approach effectively provides model ensembling benefits without requiring multiple models or additional training.

\paragraph{Adaptive Patch-Based Processing}

To handle high-resolution images efficiently, we implement an adaptive patch-based processing scheme that dynamically selects appropriate patch sizes based on input dimensions. Algorithm~\ref{alg:adaptive_inference} details our complete inference procedure.

\begin{algorithm}[t]
\caption{Adaptive Geometric Self-Ensemble Inference}
\label{alg:adaptive_inference}
\footnotesize
\begin{algorithmic}[1]
\REQUIRE Noisy image $\mathbf{X}_{t_{\text{data}}}$, model $\mathbf{h}_{\theta^*}$
\ENSURE Denoised image $\hat{\mathbf{X}}$

\STATE $\mathcal{T} \gets$ \{Identity, HFlip, VFlip, Rot90, ...\} \hfill \textcolor{gray}{$\triangleright$ 8 transforms}
\STATE $H, W \gets \text{dimensions of } \mathbf{X}_{t_{\text{data}}}$

\STATE $t_{\text{data}} \gets \begin{cases}
\text{estimate\_noise}(\mathbf{X}_{t_{\text{data}}}) & \text{if auto mode} \\
\text{predefined level} & \text{otherwise}
\end{cases}$

\STATE $\text{patch\_size} \gets \begin{cases}
896 & \text{if } \min(H, W) \geq 896 \\
768 & \text{if } \min(H, W) \geq 768 \\
512 & \text{if } \min(H, W) \geq 512 \\
\end{cases}$

\STATE $\text{stride} \gets \text{patch\_size} / 2$ \hfill \textcolor{gray}{$\triangleright$ 50\% overlap}
\STATE $\text{outputs} \gets \emptyset$

\FORALL{$T \in \mathcal{T}$}
    \STATE $\mathbf{X}_T \gets T(\mathbf{X}_{t_{\text{data}}})$
    \STATE $H_T, W_T \gets \text{dimensions of } \mathbf{X}_T$
    
    \IF{$\max(H_T, W_T) > \text{patch\_size}$}
        \STATE $\text{output}_T, \text{count} \gets \text{zeros}(H_T, W_T)$
        \STATE Pad $\mathbf{X}_T$ to dimensions divisible by stride
        
        \FOR{$(i,j)$ in overlapping patch grid}
            \STATE $\text{patch} \gets \mathbf{X}_T[i:i+\text{patch\_size}, j:j+\text{patch\_size}]$
            \STATE $\text{result} \gets \mathbf{h}_{\theta^*}(\text{patch}, t_{\text{data}})$
            \STATE Accumulate result and increment count at positions $(i,j)$
        \ENDFOR
        
        \STATE $\text{denoised}_T \gets \text{output}_T / \text{count}$
    \ELSE
        \STATE $\text{denoised}_T \gets \mathbf{h}_{\theta^*}(\mathbf{X}_T, t_{\text{data}})$
    \ENDIF
    
    \STATE $\text{outputs} \gets \text{outputs} \cup \{T^{-1}(\text{denoised}_T)\}$
\ENDFOR

\RETURN $\hat{\mathbf{X}} \gets \frac{1}{|\mathcal{T}|}\sum_{\text{out} \in \text{outputs}} \text{out}$
\end{algorithmic}
\end{algorithm}

Our adaptive patch-based approach dynamically selects from three patch sizes ($896 \times 896$, $768 \times 768$, or $512 \times 512$) based on input image dimensions. For each geometric transformation, the algorithm determines whether patch-based processing is necessary. If so, it divides the image into overlapping patches with 50\% stride, processes each patch independently, and reconstructs the full image by averaging overlapping regions. This strategy effectively handles high-resolution images while maintaining computational efficiency.

\subsection{KLETech-CEVI}
\begin{figure*}[h]
    \centering
    \includegraphics[width=1\linewidth]{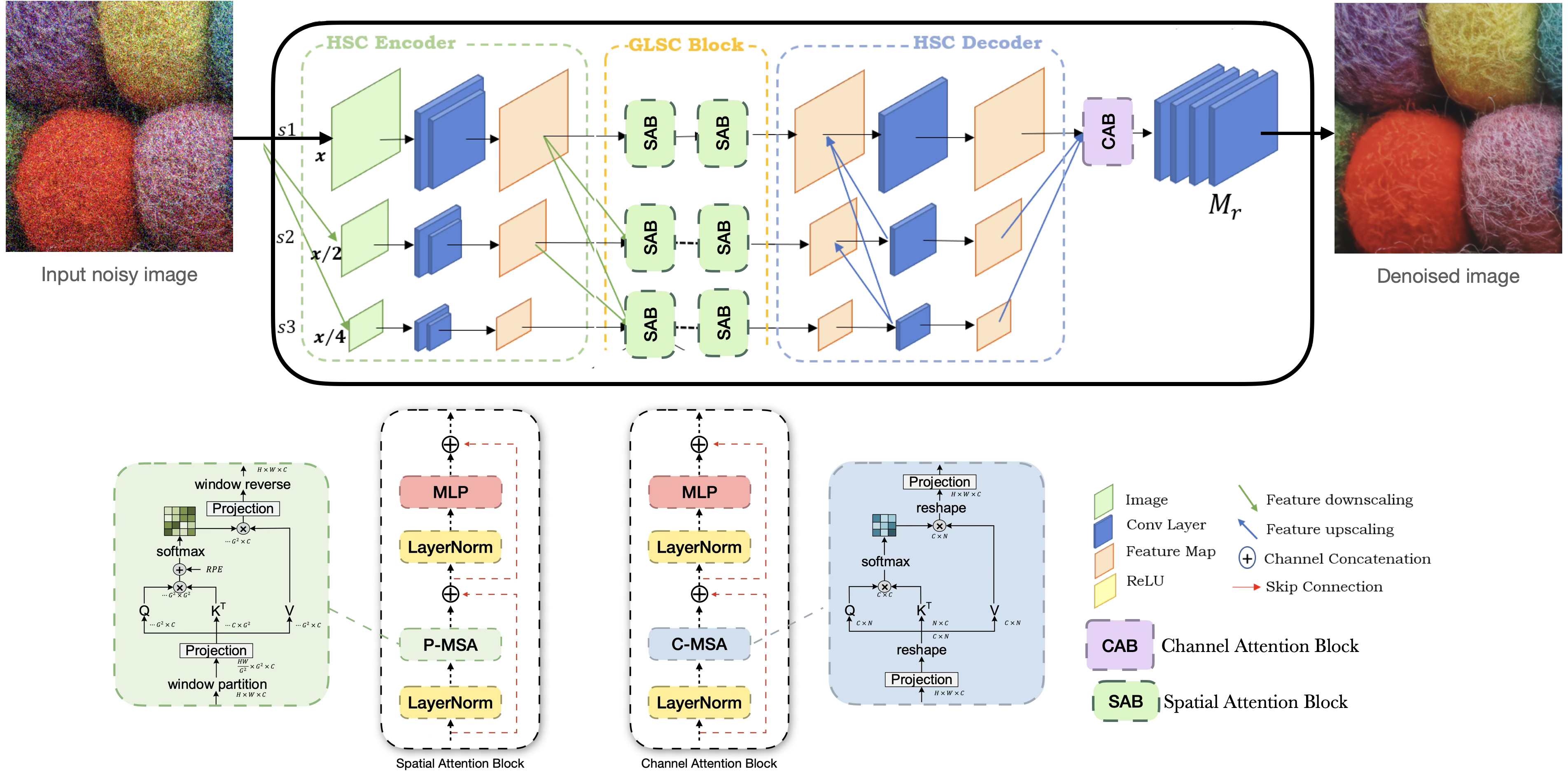}
    \caption{Overview of the HNNFormer proposed by Team KLETech-CEVI: Hierarchical Noise-Deinterlace Transformer for Image Denoising (HNNFormer). The encoder extracts features in three distinct scales, with information passed across hierarchies (green dashed box). Fine-grained global-local spatial and contextual information is learnt through the attention blocks at GLSC  (orange dashed box). At the decoder, information exchange occurs in reverse hierarchies (blue dashed box).}
    \label{fig:Methodology}
\end{figure*}

\textbf{Method: }The proposed HNNformer method is based on the HNN framework \cite{joshi2024hnn}, which includes three main modules: the hierarchical spatio-contextual (HSC) feature encoder, Global-Local Spatio-Contextual (GLSC) block, and hierarchical spatio-contextual (HSC) decoder, as shown in Figure \ref{fig:Methodology}. Typically, image denoising networks employ feature scaling for varying the sizes of the receptive fields. The varying receptive fields facilitate learning of local-to-global variances in the features. With this motivation, they learn contextual information from multi-scale features while preserving high-resolution spatial details. They achieve this via a hierarchical style encoder-decoder network with residual blocks as the backbone for learning. Given an input noisy image $x$, the proposed multi-scale hierarchical encoder extracts shallow features in three distinct scales and is given as:r

\begin{equation}
F_{si} = ME_s(x)
\end{equation}

where $F_{si}$ are the shallow features extracted at the $i^{th}$ scale from the sampled space of input noisy image $x$ and $ME_s$ represents the hierarchical encoder at scale $s$.

Inspired by \cite{zhang2023xformer}, they propose Global-Local Spatio-Contextual (GLSC) Block, that uses Spatial Attention Blocks (SAB) to learn spatial features at each scale. They also employ a Channel Attention Block (CAB) to fuse the multi-level features. The learned deep features are represented as:
\begin{equation}
D_{si} = GLSC_{si}(F_{si})
\end{equation}

where $D_{si}$ is the deep feature at the $i^{th}$ scale, $F_{si}$ are the spatial features extracted at the $i^{th}$ scale, and $GLSC_{si}$ represents Spatial Attention Blocks (SAB) at respective scales. They decode the deep features obtained at various scales with the proposed hierarchical decoder, given by:

\begin{equation}
d_{si} = MD_{si}(D_{si})
\end{equation}

where $D_{si}$ is the deep feature at the $i^{th}$ scale, $d_{si}$ is the decoded feature at the $i^{th}$ scale, and $MD_{si}$ represents the hierarchical decoder. The decoded features and upscaled features at each scale are passed to the reconstruction layers $M_r$ to obtain the denoised image $\hat{y}$. The upscaled features from each scale are stacked and represented as:

\begin{equation}
P = d_{s1} + d_{s2} + d_{s3}
\end{equation}
where $d_{s1}$, $d_{s2}$, and $d_{s3}$ are decoded features at three distinct scales, and $P$ represents the final set of features passed to the Channel Attention Block (CAB) to obtain the denoised image $\hat{y}$.

\begin{equation}
\hat{y} = M_r (P)
\end{equation}

where $\hat{y}$ is the denoised image obtained from reconstruction layers $M_r$. They optimize the learning of HNNFormer with the proposed $L_{HNNformer}$, given as:

\begin{equation}
L_{HNNformer} = (\alpha \cdot L_1) + (\beta \cdot L_{VGG}) + (\gamma \cdot L_{MSSSIM})
\end{equation}

where $\alpha$, $\beta$, and $\gamma$ are the weights. They experimentally set the weights to $\alpha = 0.5$, $\beta = 0.7$, and $\gamma = 0.5$. $L_{HNN}$ is a weighted combination of three distinct losses: $L_1$ loss to minimize error at the pixel level, perceptual loss to efficiently restore contextual information between the ground-truth image and the output denoised image, and multiscale structural dissimilarity loss to restore structural details. The aim here is to minimize the weighted combinational loss $L_{HNN}$ given as:

\begin{equation}
L(\theta) = \frac{1}{N} \sum_{i=1}^{N} \| HNNFormer(x_i) - y_i \| L_{HNN}
\end{equation}

where $\theta$ denotes the learnable parameters of the proposed framework, $N$ is the total number of training pairs, $x$ and $y$ are the input noisy and output denoised images, respectively, and $HNNFormer(\cdot)$ is the proposed framework for image denoising.
\subsection{xd\_denoise}
\begin{figure*}[!h]
\includegraphics[width=1.0\linewidth]{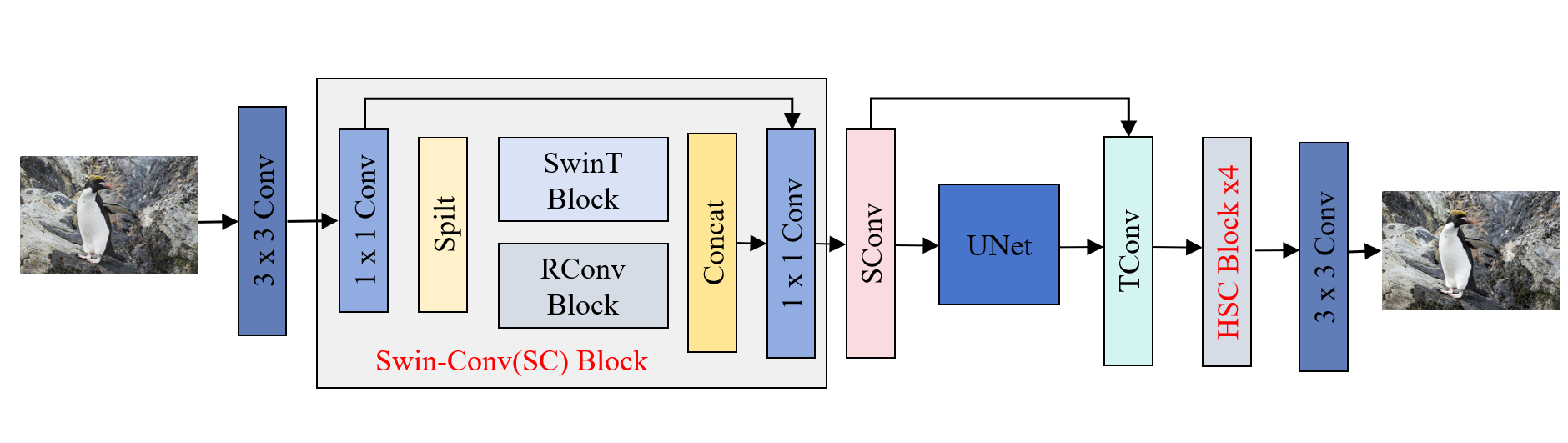}
\caption{The SCUNet model architecture proposed by Team xd\_denoise.}
\label{fig:SCUNet}
\end{figure*}

\textbf{Implementation details.} As shown in Figure~\ref{fig:SCUNet}, They use SCUNet\cite{zhang2023practical} as their baseline model.
They employed the PyTorch deep learning framework and conducted experiments on an Ubuntu 20.04 system. The hardware and software setup is as follows: CPU: Intel Xeon Gold 6226R, GPU: Four graphics cards of NVIDIA GeForce RTX 4090, Python version: 3.8.0, PyTorch version: 2.0.0, CUDA version: 11.7. They only use high-definition images from the DIV2K and LSDIR datasets for training and validation. The training set consists of 85791 images (84991 + 800), and the validation set consists of 350 images (250 + 100).
They used the Adam optimizer with 100 training epochs, a batch size of 32, and a crop size of 256 × 256. The initial learning rate was set to ${1e^{-4}}$, with ${\beta_1=0.9}$ , ${\beta_2=0.999}$, and no weight decay applied. At epoch 90, the learning rate was reduced to ${1e^{-5}}$. No data augmentation was applied during training or validation.The model is trained with MSE loss. 

\textbf{Testing description}
They integrate Test-Time Augmentation(TTA) into their method during testing, including horizontal flip, vertical flip, and 90-degree rotation. They utilized an ensemble technique by chaining three basic U-Net networks and SCUNet, and according to the weights of 0.6 and 0.4, output the results of concatenating the SCUNet model with three UNet models to achieve better performance.

\subsection{JNU620}
\begin{figure}[htbp]
    \centering
    \includegraphics[width=0.48\textwidth]{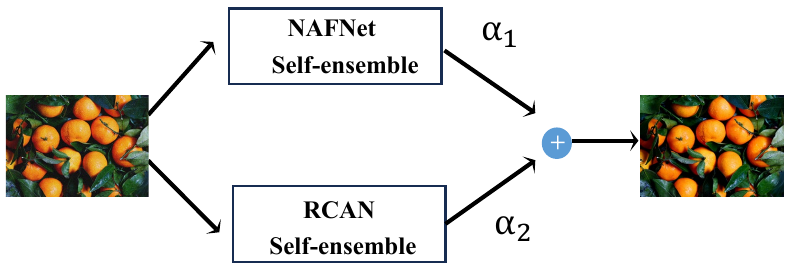}
    \caption{The pipeline of the NRDenoising proposed by Team JNU620.}
    \label{01network}
    \vspace{-.2mm}
\end{figure}

\textbf{Description.}
Recently, some research in low-level vision has shown that ensemble learning can significantly improve model performance. Thus, instead of designing a new architecture, they leverage existing NAFNet~\cite{nafnet} and RCAN~\cite{zhang2018image} as basic networks to design a pipeline for image denoising (NRDenoising) based on the idea of ensemble learning, as shown in Fig~\ref{01network}. They find the results are better improved by employing both self-ensemble and model ensemble strategies.

\textbf{Implementation details.}
For the training of NAFNet~\cite{nafnet}, they utilize the provided DIV2K~\cite{agustsson2017ntire} dataset.  The model is trained with MSE loss. They utilize the AdamW optimizer ($\beta_1=0.9$, $\beta_2=0.9$) for 400K iterations on an NVIDIA Tesla V100 GPU. The initial learning rate is set to $1\times10^{-3}$ and gradually reduces to  $1\times10^{-7}$ with the cosine annealing. The training batch is set to 4 and the patch size is 384×384. Random horizontal flipping and rotation are adopted for data augmentation.

For the training of RCAN~\cite{zhang2018image}, the provided DIV2K~\cite{agustsson2017ntire} dataset is also employed. The MSE loss is utilized with an initial learning rate of $1\times10^{-4}$. The Adam optimizer ($\beta_1=0.9$, $\beta_2=0.99$) is used for 100K iterations. The batch size is 3, and the patch size is 200×200. Data augmentation includes the horizontal flip and the 90-degree rotation.

During inference, they apply a self-ensemble strategy for NAFNet~\cite{nafnet} and selectively adopt the TLC~\cite{chu2022improving} method based on the size of input images; For RCAN~\cite{zhang2018image}, they utilize a self-ensemble strategy. Finally, the model-ensemble strategy is employed to combine the outputs of NAFNet~\cite{nafnet} and RCAN~\cite{zhang2018image}.

\subsection{PSU\_team}

\begin{figure*}[t]
\centering
\includegraphics[width=0.99\textwidth]{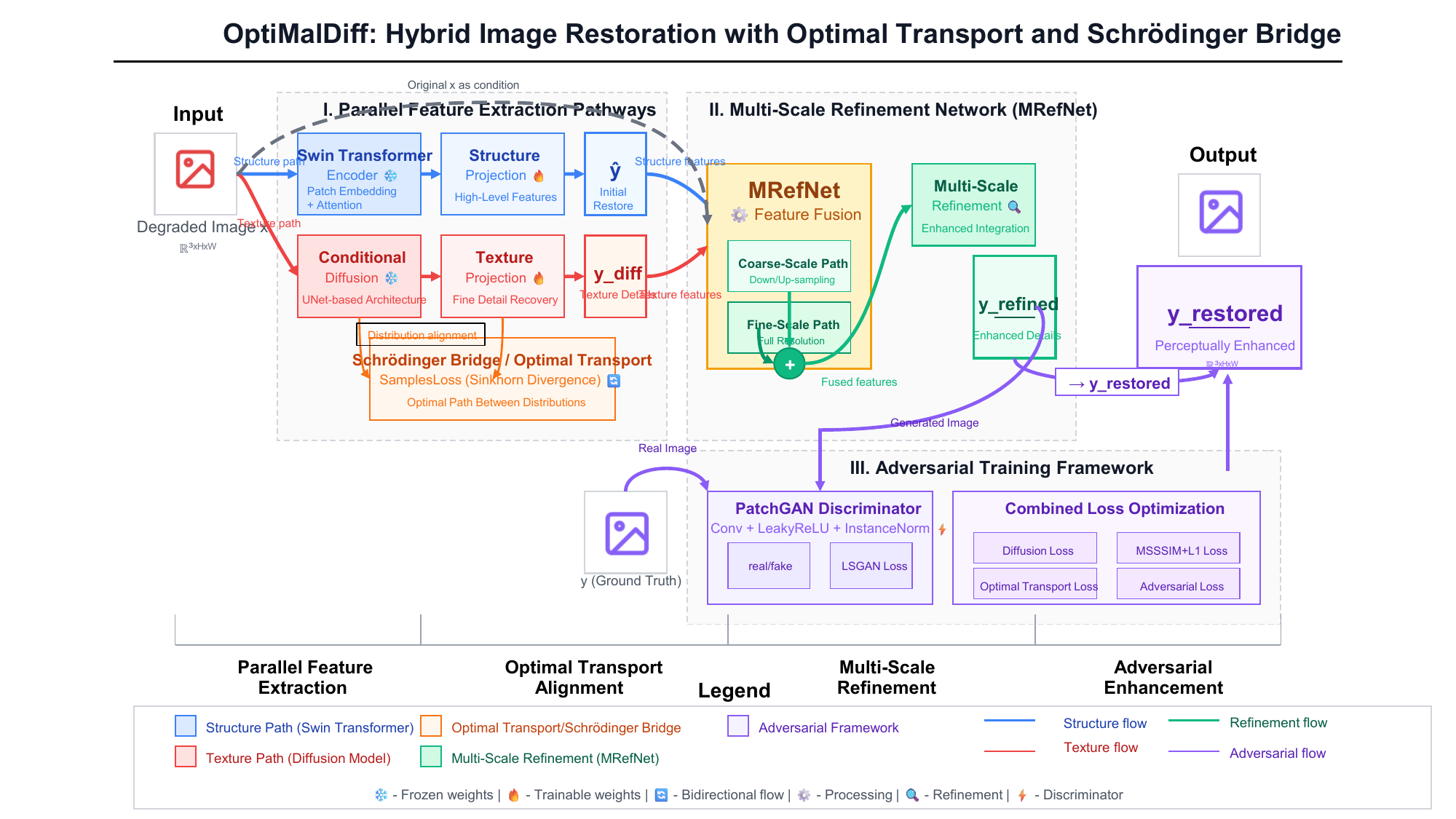}
\caption{Overview of the OptiMalDiff architecture proposed by PSU\_team, combining Schrödinger Bridge diffusion, transformer-based feature extraction, and adversarial refinement.}
\label{fig:method_psu}
\end{figure*}

\textbf{General method description.} They propose \textbf{OptiMalDiff}, a high-fidelity image enhancement framework that reformulates image denoising as an optimal transport problem. The core idea is to model the transition from noisy to clean image distributions via a Schrödinger Bridge-based diffusion process. The architecture (shown in Fig.~\ref{fig:method_psu}) consists of three main components: (1) a hierarchical Swin Transformer backbone that extracts both local and global features efficiently, (2) a Schrödinger Bridge Diffusion Module that learns forward and reverse stochastic mappings, and (3) a Multi-Scale Refinement Network (MRefNet) designed to progressively refine image details. To enhance realism, they integrate a PatchGAN discriminator with adversarial training.

\textbf{Training details.} The model is trained from scratch using the DIV2K dataset, without relying on any pre-trained weights. They jointly optimize all modules using a composite loss function that includes diffusion loss, Sinkhorn-based optimal transport loss, multi-scale SSIM and L1 losses, and an adversarial loss. The training spans 300 epochs with a batch size of 8, totaling 35,500 iterations per epoch. The method emphasizes both fidelity and perceptual quality, achieving strong results in PSNR and LPIPS.

\subsection{Aurora}

They will introduce their algorithm from four aspects: model architecture, data processing methods, training pipeline, and testing pipeline.

Given the excellent performance of generative adversarial networks (GANs) in image generation tasks, and considering that image denoising can also be regarded as a type of generative task, they utilize a generative adversarial network for the denoising task. Specifically, they adopt NAFNet \cite{nafnet} as the generator and have made a series of parameter adjustments. In particular, they increased both the number of channels and the number of modules. Due to the superior performance of the SiLU activation function across various tasks, they replaced the original activation function with SiLU. For the discriminator, they employ a VGG11 architecture without batch normalization (BN) layers, where the ReLU activation function is replaced with LeakyReLU.

In the training stage, they exclusively use the DIV2K and LSDIR datasets \cite{lilsdir}. Instead of employing overly complex data augmentation algorithms, they applied simple flipping and rotation techniques for data augmentation. Finally, a patch is cropped from the high-resolution (HR) image, normalized, and then fed into the network.

During training, they progressively trained the model using resolutions of [128, 192, 256]. The model was jointly optimized using L1, L2, and Sobel loss functions. The optimizer and learning rate scheduler used during training were AdamW and CosineAnnealingLR, respectively.

In the inference phase, they employed a self-ensemble strategy and selectively adopted the TLC \cite{chu2021revisiting} method to further enhance performance.
\subsection{mpu\_ai}

\subsubsection{Method}
\begin{figure*}[tbp]
  \centering
  \includegraphics[width=\textwidth]{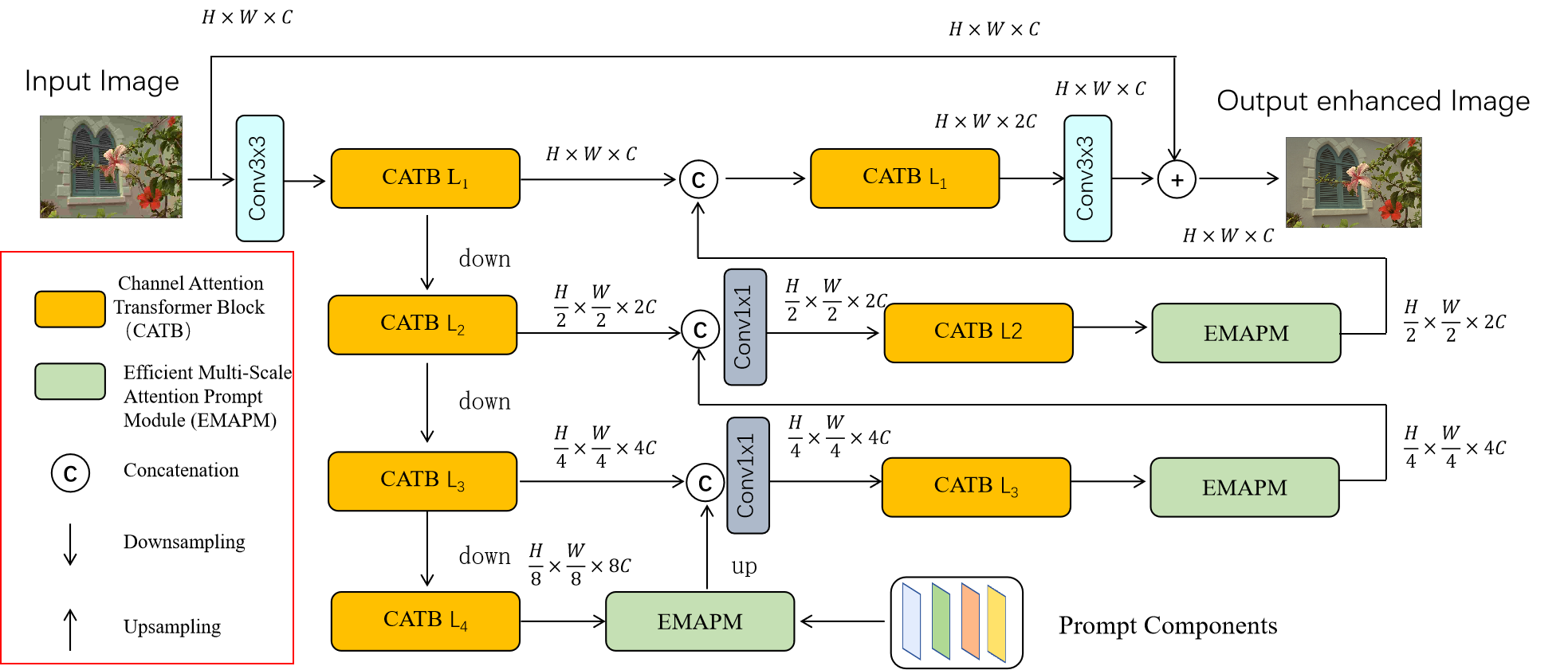}
  \caption{The CTMP architecture proposed by Team mpu\_ai}
  \label{mpuai_fig1}
\end{figure*}

Existing deep learning-based image restoration methods exhibit inadequate generalization capabilities when faced with a variety of noise types and intensities, thereby significantly impeding their broad application in real-world scenarios. To tackle this challenge, this paper proposes a novel prompt-based learning approach, namely Blind Image Restoration Using Dual-Channel Transformers and Multi-Scale Attention Prompt Learning (CTMP), as depicted in Figure~\ref{mpuai_fig1}. The CTMP model features a U-shaped architecture grounded in the Transformer framework, constructed from the enhanced Channel Attention Transformer Block (CATB). During the image restoration process, CTMP adopts a blind image restoration strategy to address diverse noise types and intensities. It integrates an Efficient Multi-Scale Attention Prompt Module (EMAPM) that is based on prompts. Within the EMAPM, an Enhanced Multi-scale Attention (EMA) module is specifically designed. This module extracts global information across different directions and employs dynamic weight calculations to adaptively modulate the importance of features at various scales. The EMA module subsequently fuses the enhanced multi-scale features with the input feature maps, yielding a more enriched feature representation. This fusion mechanism empowers the model to more effectively capture and leverage features at different scales, thereby markedly bolstering its capacity to restore image degradations and showcasing superior generalization capabilities.

\begin{figure*}[t]
  \centering
  \includegraphics[width=\textwidth]{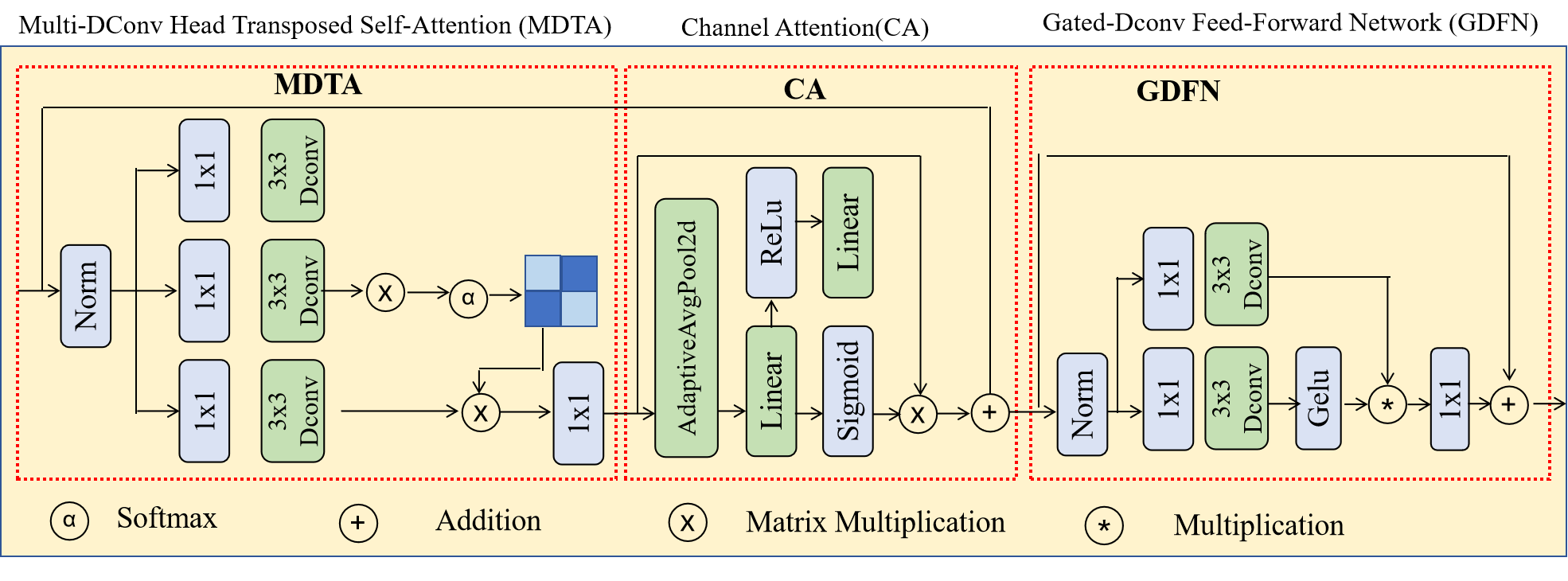}
  \caption{The Channel Attention Transformer Block (CATB), proposed by Team mpu\_ai}
  \label{mpuai_fig2}
\end{figure*}

\subsubsection{Transformer Block Incorporating Channel Attention and Residual Connections}

The Transformer Block serves as the cornerstone of their entire model, harnessing the Transformer architecture to extract image features through the self-attention mechanism. In pursuit of enhanced performance, they have refined the Transformer module by devising a novel architecture that integrates Channel Attention with the self-attention mechanism, thereby combining the strengths of both Transformer and Channel Attention. Specifically, the Transformer focuses on extracting high-frequency information to capture the fine details and textures of images, while Channel Attention excels at capturing low-frequency information to extract the overall structure and semantic information of images. This integration further boosts the image denoising effect.As depicted in Figure~\ref{mpuai_fig2}, the improved Transformer architecture, named the Channel Attention Transformer Block (CATB), primarily consists of the following three modules: Multi-DConv Head Transposed Self-Attention (MDTA), Channel Attention (CA), and Gated-Dconv Feed-Forward Network (GDFN). 

The Multi-DConv Head Transposed Self-Attention (MDTA) module enhances the self-attention mechanism's perception of local image features by incorporating multi-scale depthwise convolution operations, effectively capturing detailed image information. The Channel Attention (CA) module, dedicated to information processing along the channel dimension, computes the importance weights of each channel to perform weighted fusion of channel features, thereby strengthening the model's perception of the overall image structure. The Gated-Dconv Feed-Forward Network (GDFN) module combines the gating mechanism with depthwise convolution operations, aiming to further optimize the nonlinear transformation of features. By introducing the gating mechanism, the model can adaptively adjust the transmission and updating of features based on the dynamic characteristics of the input features, thereby enhancing the flexibility and adaptability of feature representation. Through the synergistic action of these three modules, the improved Transformer architecture can more effectively handle both high-frequency and low-frequency information in images, thereby significantly enhancing the performance of image denoising and restoration.

In image restoration tasks, feature extraction and representation are crucial steps. Traditional convolutional neural networks (CNNs) and Transformer architectures primarily focus on feature extraction in the spatial domain, while paying less attention to the weighting of features in the channel dimension. To address this limitation, they introduce a Channel Attention module in the Transformer Block, creating a Transformer Block that incorporates Channel Attention and Residual Connections. This module weights the channel dimension through global average pooling and fully connected layers, enhancing important channel features while suppressing less important ones. This weighting mechanism enables the model to focus more effectively on key information, thereby improving the quality of restored images. Additionally, the introduction of residual connections further enhances the model's robustness and performance. Residual connections ensure that the information of the input features is fully retained after processing by the Channel Attention module by adding the input features directly to the output features. This design not only aids gradient propagation but also retains the original information of the input features when the weighting effect of the Channel Attention module is suboptimal, further boosting the model's robustness.

The proposed model incorporates several key enhancements to improve image restoration quality. Firstly, the Channel Attention Module leverages global average pooling and fully connected layers to selectively enhance important channel features while suppressing less relevant ones. This mechanism enables the model to focus more effectively on critical information, thereby improving the quality of the restored image. Secondly, residual connections are employed to ensure that the original input features are fully retained and added directly to the output features after processing by the Channel Attention Module. This not only aids gradient propagation but also preserves the original information when the weighting effect is suboptimal, thus boosting the model's robustness. Lastly, the LeakyReLU activation function is utilized in the Feed-Forward Network to introduce non-linearity while avoiding the "dying neurons" issue associated with ReLU, further enhancing the model's expressive power. Together, these improvements contribute to a more effective and robust image restoration model.

\subsubsection{Efficient Multi-Scale Attention Prompt Module}
\begin{figure*}[htbp]
  \centering
  \includegraphics[width=\textwidth]{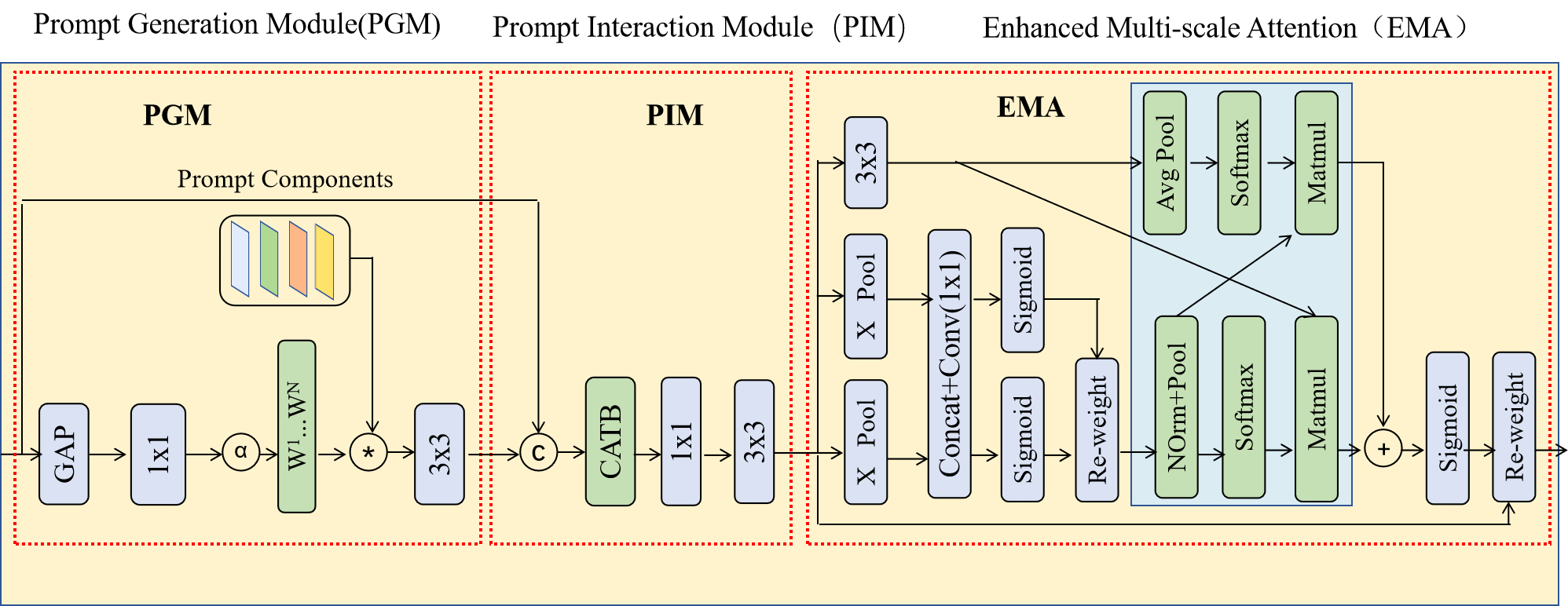}
  \caption{Efficient Multi-Scale Attention Prompt Module (EMAPM), proposed by Team mpu\_ai.}
  \label{mpuai_fig3}
\end{figure*}

Addressing multi-scale image degradations is a crucial challenge in image restoration tasks. Traditional feature extraction methods typically capture features at a single scale, neglecting the fusion and interaction of features across multiple scales. To overcome this limitation, they propose a prompt-based blind image restoration approach, incorporating an Efficient Multi-Scale Attention Prompt Module (EMAPM). As be shown in Figure~\ref{mpuai_fig3}, the core of the EMAPM is the Enhanced Multi-scale Attention (EMA) module, which extracts global information in different directions and combines dynamic weight calculations to adaptively adjust the significance of features at various scales, thereby generating a richer feature representation. This design not only enhances the model's adaptability to multi-scale image degradations but also strengthens the expressiveness of features, significantly improving the quality of image restoration. The introduction of the EMA module represents a significant innovation in their image restoration approach. Experimental results validate the effectiveness of the EMA module, demonstrating its ability to substantially boost model performance across multiple image restoration tasks. This innovation not only enhances the model's restoration capabilities but also offers new research directions for image restoration tasks.

The Efficient Multi-Scale Attention Prompt Module (EMAPM) is designed to enhance the model's ability to capture multi-scale features in image restoration tasks. By generating adaptive prompts that focus on different scales and characteristics of the input image, EMAPM allows the model to better handle various types of image degradations. The core components and operations of EMAPM are described as follows:

\textbf{Module Configuration:} To configure the EMAPM, several key parameters are defined:
\begin{itemize}
    \item \textbf{Prompt Dimension ($d_p$)}: This determines the dimension of each prompt vector, which represents the feature space for each prompt.
    \item \textbf{Prompt Length ($L_p$)}: This specifies the number of prompt vectors, which controls the diversity of prompts generated.
    \item \textbf{Prompt Size ($S_p$)}: This sets the spatial size of each prompt vector, which affects the resolution of the prompts.
    \item \textbf{Linear Dimension ($d_l$)}: This is the dimension of the input to the linear layer, which processes the embedding of the input feature map.
    \item \textbf{Factor ($f$)}: This defines the number of groups in the EMA module, which influences the grouping mechanism in the attention process.
\end{itemize}

\textbf{Mathematical Formulation:} Given an input feature map \( x \in \mathbb{R}^{B \times C \times H \times W} \), where \( B \) is the batch size, \( C \) is the number of channels, and \( H \times W \) is the spatial dimension, the operations within EMAPM are defined as follows:

1. \textbf{Compute Embedding:} The embedding of the input feature map is computed by averaging the spatial dimensions.
   \begin{equation}
       \text{emb} = \frac{1}{H \times W} \sum_{i=1}^{H} \sum_{j=1}^{W} x_{:, :, i, j} \in \mathbb{R}^{B \times C}
   \end{equation}

2. \textbf{Linear Layer and Softmax:} The embedding is passed through a linear layer followed by a softmax function to generate prompt weights.
   \begin{equation}
       \text{prompt\_weights} = \text{softmax}(\text{linear\_layer}(\text{emb})) \in \mathbb{R}^{B \times L_p}
   \end{equation}

3. \textbf{Generate Prompt:} The prompts are generated by weighting the prompt parameters with the prompt weights and then summing them up. The prompts are then interpolated to match the spatial dimensions of the input feature map.
   \begin{equation}
       \text{prompt} = \sum_{k=1}^{L_p} \text{prompt\_weights}_{:, k} \cdot \text{prompt\_param}_{k} \in \mathbb{R}^{B \times d_p \times S_p \times S_p}
   \end{equation}
   \begin{equation}
       \text{prompt} = \text{F.interpolate}(\text{prompt}, (H, W), \text{mode}=\text{"bilinear"})
   \end{equation}

4. \textbf{Enhance Prompt using EMA:} The prompts are enhanced using the Enhanced Multi-scale Attention (EMA) module, which refines the prompts by incorporating multi-scale attention.
   \begin{equation}
       \text{enhanced\_prompt} = \text{EMA}(\text{prompt}) \in \mathbb{R}^{B \times d_p \times H \times W}
   \end{equation}

5. \textbf{Conv3x3:} Finally, the enhanced prompts are processed through a 3x3 convolutional layer to further refine the feature representation.
   \begin{equation}
       \text{enhanced\_prompt} = \text{conv3x3}(\text{enhanced\_prompt}) \in \mathbb{R}^{B \times d_p \times H \times W}
   \end{equation}

\subsubsection{Experiments}
In this section, they conducted a series of extensive experiments to comprehensively demonstrate the superior performance of the proposed CTMP model across multiple datasets and benchmarks. The experiments covered a variety of tasks, including denoising and deblocking of compressed images, and were compared with previous state-of-the-art methods. Additionally, they reported the results of ablation studies, which strongly validated the effectiveness of the Channel Attention Transformer Block (CATB) and the Enhanced Multi-scale Attention Prompt Module (EMAPM) within the CTMP architecture.

The CTMP framework is end-to-end trainable without the need for pretraining any individual components. Its architecture consists of a 4-level encoder-decoder, with each level equipped with a different number of Transformer modules, specifically [4, 6, 6, 8] from level 1 to level 4. They placed a Prompt module between every two consecutive decoder levels, resulting in a total of 3 Prompt modules across the entire PromptIR network, with a total of 5 Prompt components. During training, the model was trained with a batch size of 2, leveraging the computational power of a Tesla T4 GPU. The network was optimized through L1 loss, using the Adam optimizer ($\beta_1 = 0.9$, $\beta_2 = 0.999$) with a learning rate of $2 \times 10^{-4}$. To further enhance the model's generalization ability, they used 128×128 cropped blocks as input during training and augmented the training data by applying random horizontal and vertical flips to the input images.

The proposed model in this paper exhibits the following characteristics in terms of overall complexity: It consists of approximately \(35.92\) million parameters and has a computational cost of \(158.41\) billion floating-point operations (FLOPs). The number of activations is around \(1{,}863.85\) million, with \(304\) Conv2d layers. During GPU training, the maximum memory consumption is \(441.57\) MB, and the average runtime for validation is \(25{,}287.67\) seconds.

\subsubsection{Dataset}
To comprehensively evaluate the performance of the CTMP algorithm in image restoration tasks, they conducted experiments in two critical areas: image denoising and deblocking of compressed images. For training, they selected the high-quality DIV2K dataset, which comprises 800 high-resolution clean images with rich textures and details, providing ample training samples to enable the model to perform well under various degradation conditions \citep{agustsson2017ntire}. Additionally, they used 100 clean/noisy image pairs as the validation set to monitor the model's performance during training and adjust the hyperparameters.

During the testing phase, they chose several widely used datasets, including Kodak, LIVE1, and BSDS100, to comprehensively assess the algorithm's performance. The Kodak dataset consists of 24 high-quality images with diverse scenes and textures, commonly used to evaluate the visual effects of image restoration algorithms \citep{kodak}. The LIVE1 dataset contains a variety of image types and is widely used for image quality assessment tasks, effectively testing the algorithm's performance under different degradation conditions \citep{live1}. The BSDS100 dataset includes 100 images with rich textures and edge information, providing a comprehensive evaluation of the algorithm's performance in image restoration tasks \citep{bsds100}.

By testing on these representative datasets, they were able to comprehensively evaluate the CTMP algorithm's performance across different degradation types and image conditions, ensuring its effectiveness and reliability in practical applications.

\subsection{OptDenoiser}
\begin{figure}
  \includegraphics[width=0.5\textwidth]{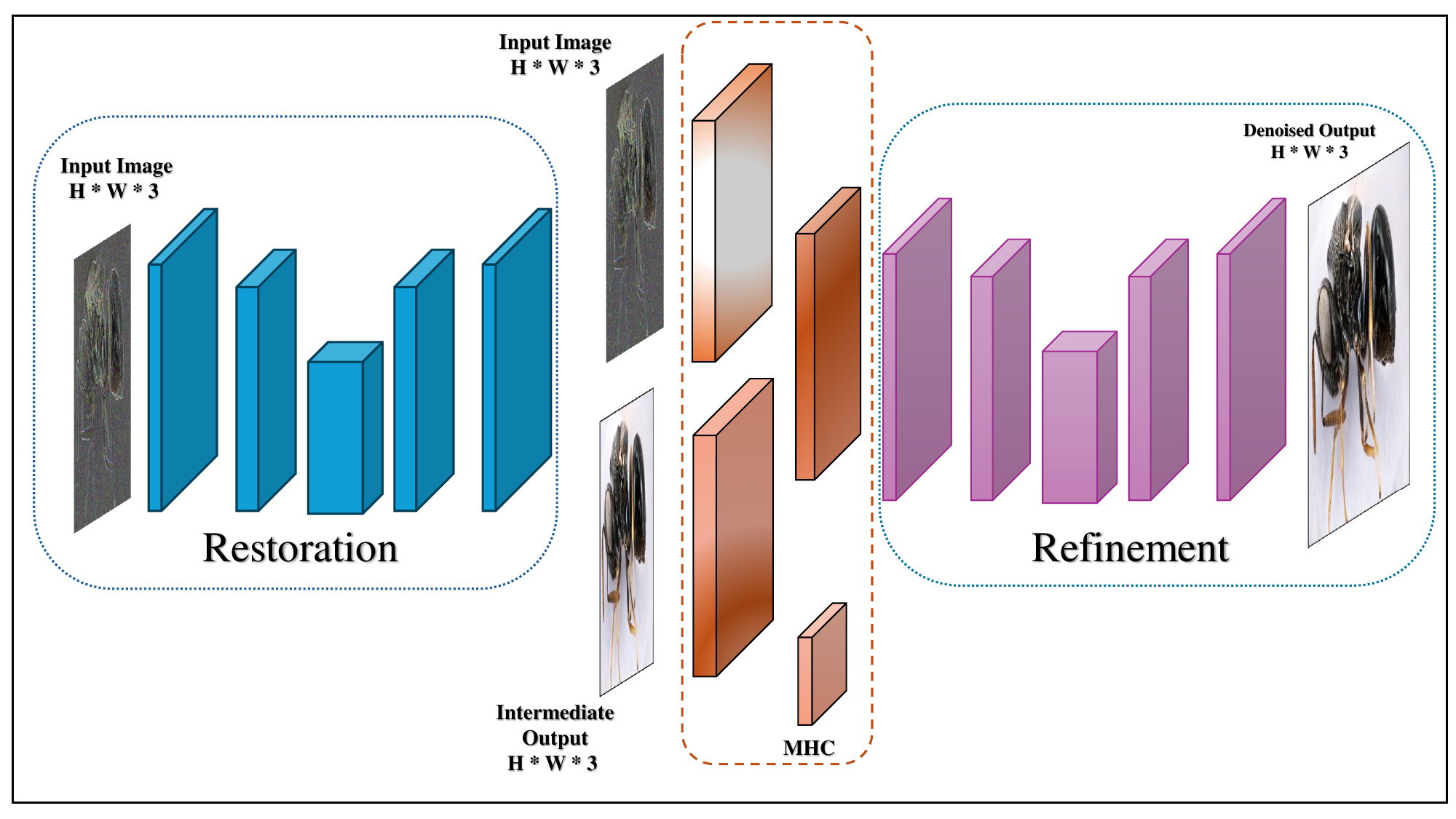}
\caption{Overview of the two-stage OptDenoiser framework for image denoising.}
  \label{opt}
\end{figure}

\textbf{Method} They introduce a two-stage transformer-based network that effectively maps low-resolution noisy images to their high-resolution counterparts, as depicted in Fig. \ref{opt}. The proposed framework comprises two independent encoder-decoder blocks (EDBs) and  Multi-Head correlation blocks to generate visually coherent images \cite{sharif2025illuminating}. To enhance reconstruction efficiency, they integrate illumination mapping \cite{sharif2025illuminating} guided by Retinex theory \cite{land1971lightness}. Additionally, they conduct a theory, an in-depth evaluation of the effectiveness of illumination mapping in general image reconstruction tasks, including image denoising. Therefore, their framework integrates the Retinexformer \cite{cai2023retinexformer} network as the first stage. In the context of image denoising, Retinexformer surpasses conventional denoisers such as UFormer, Restormer, and DnCNN. However, like other denoising methods, Retinexformer encounters challenges, including jagged edges, blurred outputs, and difficulties in capturing and representing complex structures in noisy inputs. To address these obstacles, they incorporate the MHC, followed by an additional EDB  in their framework. This design effectively exploits feature correlations from intermediate outputs, enabling more accurate reconstruction with improved structural fidelity and texture preservation. Furthermore, they integrate a perceptual loss function with luminance-chrominance guidance \cite{sharif2025illuminating} to mitigate color inconsistencies, ensuring visually coherent and perceptually refined reconstructions.
\subsubsection{Global Method Description}

\textbf{Training Procedure:} During the training phase, input images were randomly cropped into 512 × 512 patches and subsequently downscaled to 128 × 128 to enhance the model’s ability to capture spatial features effectively. A fixed learning rate of 0.0001 was maintained throughout the training process. The model was trained exclusively on the LSDIR and DIV2K datasets, without the inclusion of any additional training, validation, or testing data.
\subsubsection{Technical details}
The proposed solution is implemented with the \textit{PyTorch} framework. The networks were optimized
using the \textit{Adam} optimizer, where the hyperparameters were tuned as $\beta_1 = 0.9$, $\beta_2 = 0.99$, and the learning
rate was set to $1 \times 10^{-4}$. They trained their model using randomly cropped image patches with a constant batch size of
4, which takes approximately 72 hours to complete. All experiments were conducted on a machine equipped with an \textit{NVIDIA RTX 3090 GPU}.

\begin{figure*}
    \centering
    \includegraphics[width=1.0\linewidth]{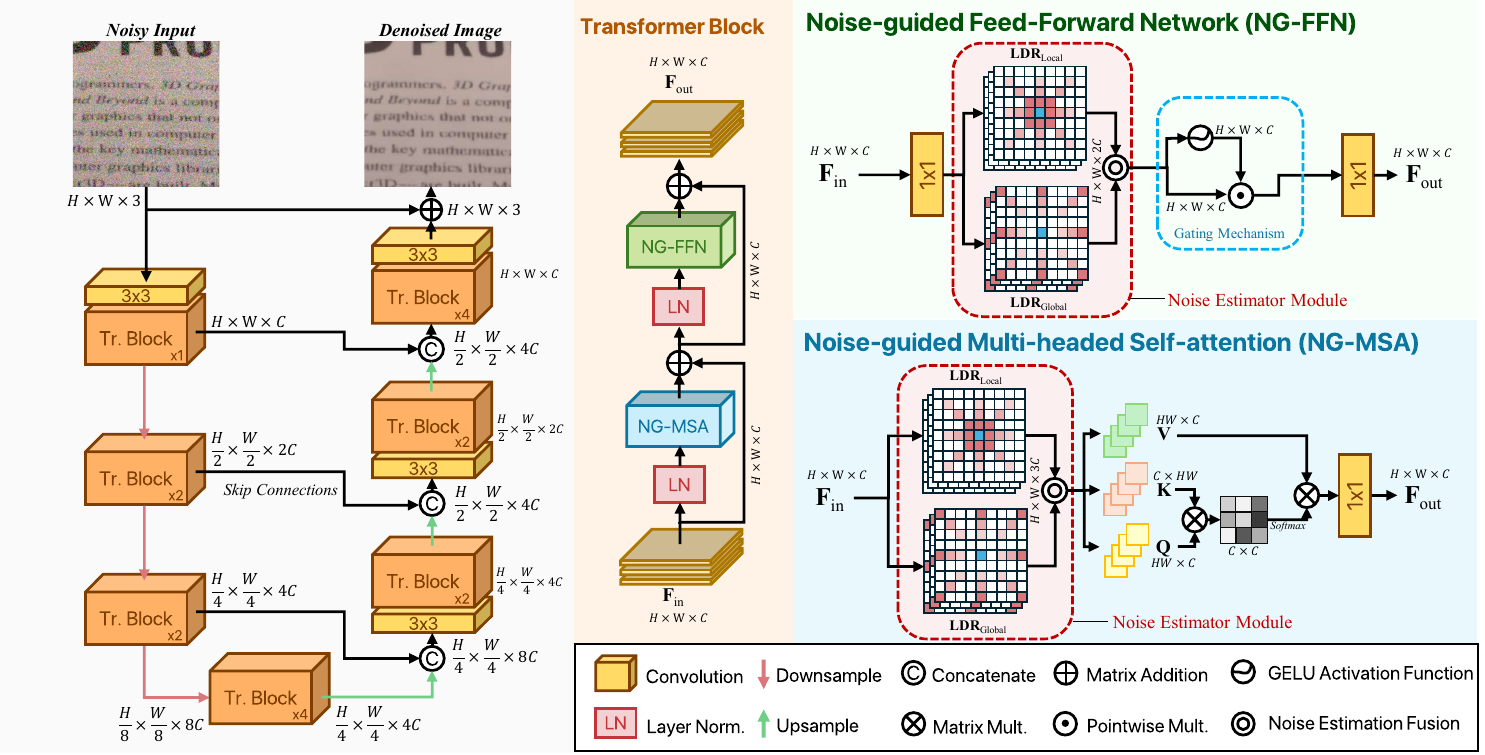}
    \caption{Overall framework of \textbf{AKDT} - Adaptive Kernel Dilation Transformer.}
    \label{fig:akdt_framework}
\end{figure*}

\subsection{AKDT}

\noindent\textbf{Method.} The team utilizes their existing network \textit{Adaptive Kernel Dilation Transformer}~\cite{brateanu418akdt} (\textbf{AKDT}), published at VISAPP 2025, with code published at \url{https://github.com/albrateanu/AKDT}. Figure~\ref{fig:akdt_framework} presents the architecture of \textbf{AKDT}. It proposes a novel convolutional structure with learnable dilation rates: the Learnable Dilation Rate (\textbf{LDR}) Block, used to formulate the Noise Estimator (\textbf{NE}) Module, which is leveraged within the self-attention and feed-forward mechanisms.

\noindent{\textbf{LDR.}} The Learnable Dilation Rate module lies at the foundation of \textbf{AKDT} and helps the model effectively pick optimal dilation rates for convolutional kernels. Given an input feature map \( \textbf{F}_\text{in} \in \mathbb{R}^{H \times W \times C} \), it is formulated as the weighted concatenaton of $N$ dilated convolutions:
    \begin{equation} 
        \label{formula:akdt_ldr}
            \textbf{F}_\text{LDR}\!=\!\textit{conv}1\!\!\times\!\!1\left( \textbf{concat} _{i=1}^{N} \alpha_i \times \textit{conv}3\!\!\times\!\!3_i(\textbf{F}_\text{in})\right)
    \end{equation}

\noindent where \textbf{concat} represents the channel-wise concatenation operation. The specific dilation rates picked for LDR are a hyperparameter that is carefully chosen to balance between performance and computational efficiency.

\noindent{\textbf{NE.}} The Noise Estimator integrates both global and local context understanding through its unique structure. This module consists of two distinct parallel components: the Global and Local \textbf{LDR} modules with selected dilation rates for capturing global and local structure. It is defined as:

    \begin{equation} 
        \label{formula:akd_ne}
            \textbf{NE}= \varrho \left( \textbf{LDR}_\text{Global}, \textbf{LDR}_\text{Local}\right)
    \end{equation}

\noindent where $\varrho$ is the Noise Estimation Fusion operation that merges global and local noiseless feature context. 

\noindent{\textbf{NG-MSA.}} To ensure efficiency in their Noise-guided Multi-headed Self-Attention, they utilize the Transposed Multi-headed Self-Attention mechanism~\cite{restormer} as baseline. They then integrate their proposed \textbf{NE} module for the \textbf{Q,K,V} extraction phase, to ensure self-attended feature maps are produced utilizing noiseless context. Therefore, given the input feature map $\textbf{F}_\text{in} \in \mathbb{R}^{H \times W \times C}$, they can define this process as:

    \begin{equation} 
        \label{formula:akdt_qkv}
            \{\textbf{Q,K,V}\}= \textbf{NE}(\textbf{F}_\text{in}), \quad \textbf{Q,K,V} \in \mathbb{R}^{HW \times C}
    \end{equation}

\noindent Then, \textbf{Q,K} are used to compute the self-attention map by matrix multiplication and Softmax activation, which is then applied to \textbf{V} to obtain the final self-attended feature map.

\noindent{\textbf{NG-FFN.}} The Noise-guided Feed-forward Network also utilizes the \textbf{NE} module for noise-free feature extraction context. It consists of a series of convolutional layers with a gating mechanism used to selectively apply non-linear activations. The noise-free features, obtained from projecting the input through their \textbf{NE} will be referred to as $\textbf{F}_\text{NE} \in \mathbb{R}^{H \times W \times C}$. Consequently, the feed-forward process can be described as:

        \begin{equation} \label{formula:NG-FFN}
           \textbf{F}_\text{NG-FFN} = \phi\left(W_1 \textbf{F}_\text{NE}\right) \odot W_2 \textbf{F}_\text{NE} + \textbf{F}_\text{NE},
        \end{equation}

\noindent here $\phi$ denotes the GELU activation function, $\odot$ represents element-wise multiplication, and $W_1$, $W_2$ are the learnable parameters of the parallel paths.

\noindent{\textbf{Implementation.}} \textbf{AKDT} is implemented by PyTorch. They only use the DIV2K dataset for training. The model is trained using the Adam Optimizer for 150k iterations, with an initial learning rate set at $2e-4$ which gradually decreases through a Cosine Annealing scheme. Each iteration consists of a batch of 4 $600\times600$ randomly-cropped image patches that undergo data augmentation (random flipping/rotation). To optimize their network, they utilize a hybrid loss function capable to capture pixel-level, multi-scale and perceptual differences~\cite{brateanu2025enhancing}~\cite{brateanu2024kolmogorov}. Testing is performed via standard inference, without additional enhancement techniques.
\subsection{X-L}
\begin{figure}[t]
\centering
\includegraphics[width=23em]{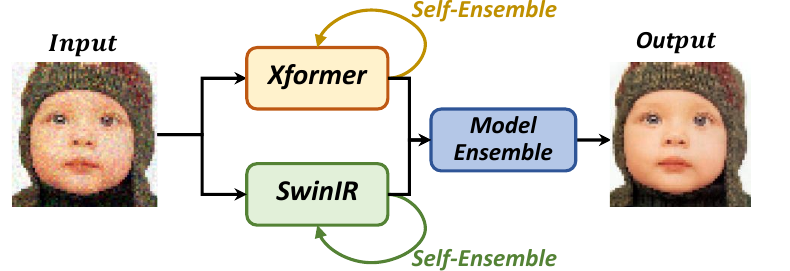}
\caption{{Overview of the MixEnsemble pipeline proposed by Team X-L.}
\label{fig:method2}}
\end{figure}

\textbf{General method description.} To ensure performance while reducing computational overhead, they adopted the following strategy: leveraging two leading approaches, Xformer~\cite{zhang2023xformer} and SwinIR~\cite{swinir}, the pipeline is shown in Fig.~\ref{fig:method2}. 
They directly utilized their pre-trained models to perform self-ensemble, generating two output results. 
Then, they conducted model ensemble on these two outputs, integrating the results between models to obtain the final reconstruction result.

\textbf{Training details.} They do not require additional training; instead, they directly leverage existing methods and their pre-trained models for inference. This approach not only saves significant computational resources and time but also fully utilizes the excellent models and valuable expertise available in the field. By directly employing these pre-trained models, they can quickly generate high-quality predictions while avoiding the high costs and complexity associated with training models from scratch.

\subsection{Whitehairbin}
\subsubsection{Introduce}
Their method is based on the Refusion\cite{luo2023refusion} model proposed in previous work, and they trained it on the dataset provided by this competition to validate its effectiveness. The Refusion model itself is a denoising method based on the diffusion model framework. Its core idea is to guide the reverse diffusion process by learning the noise gradient (score function) at different time steps t. Within the Refusion framework, they can still flexibly choose NAFNet or UNet as the neural network backbone architecture to adapt to different computational resources and performance requirements. NAFNet is known for its efficiency, while UNet excels in preserving details. The denoising process follows a stochastic differential equation (SDE) approach, which calculates the score function by predicting the noise residual and iteratively removes noise. Through training and validation on the competition dataset, their method ultimately achieved a test performance of PSNR 27.07 and SSIM 0.79.
\subsubsection{Method details}
\textbf{General method description } 
Their proposed denoising method is based on a diffusion model framework, where the network is designed to estimate the noise gradient (\textit{score function}) at different time steps $t$ to guide the reverse diffusion process. The core architecture consists of a neural backbone, which can be either NAFNet, selected based on a trade-off between computational efficiency and denoising quality.

NAFNet features a lightweight structure optimized for high-speed image restoration, incorporating a self-gated activation mechanism (SimpleGate), simplified channel attention (SCA), and depth-wise convolutions, making it highly efficient. UNet, on the other hand, is a widely adopted architecture for image denoising, leveraging an encoder-decoder structure with skip connections to preserve spatial details while extracting multi-scale features.

The denoising process follows a stochastic differential equation (SDE) approach, where Gaussian noise $\mathcal{N}(0, \sigma_t^2 I)$ is added to the clean image $x_0$ during the forward diffusion process, and the network is trained to predict the noise residual $s_\theta(x_t, t)$. This predicted noise is used to compute the \textit{score function}, which guides the reverse diffusion process, progressively removing noise through an iterative update step:
\[
x_{t-1} = x_t - 0.5 \cdot \sigma_t^2 \cdot \text{score}(x_t, t) \cdot dt.
\]

To improve sampling efficiency, they integrate an ODE-based sampling strategy, which allows for faster denoising while maintaining high restoration quality. Additionally, they employ a cosine noise schedule, which ensures a smooth noise transition across time steps and improves training stability. The network is optimized using a custom loss function that minimizes the deviation between the predicted noise and the true noise, ensuring precise score estimation.

Training is conducted with the Lion optimizer, incorporating a learning rate scheduler for improved convergence. To enhance computational efficiency, they apply mixed precision training, reduce time steps $T$, and utilize lightweight backbone networks, striking a balance between high-quality denoising and efficient execution.
\begin{figure*}[t]
    \centering
    \includegraphics[width=1\linewidth]{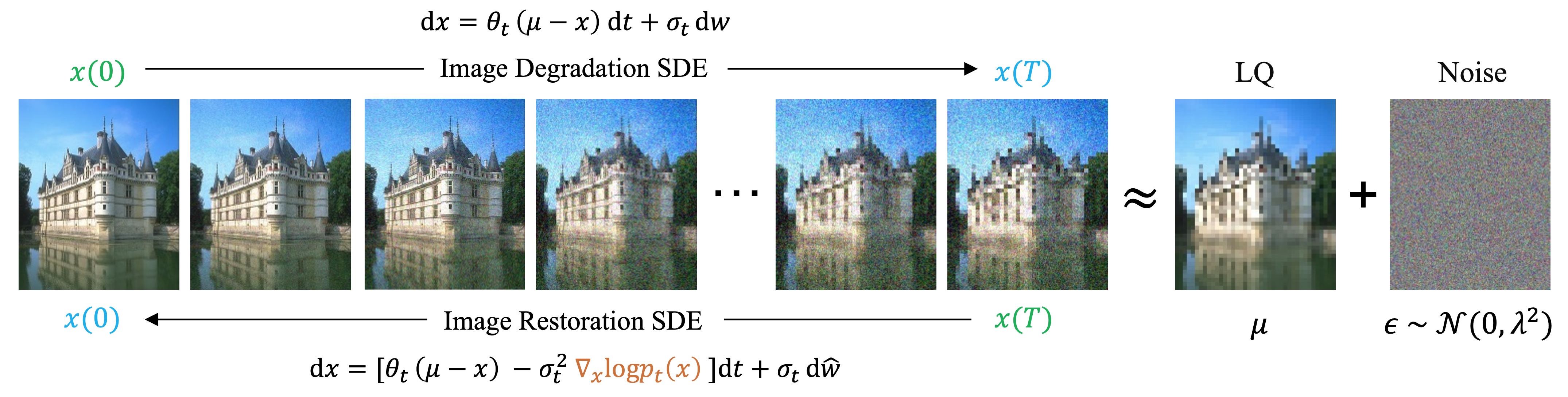} 
    \caption{Diffusion model for image denoising from Team Whitehairbin.}
    \label{fig:diffusion-model}
\end{figure*}

\textbf{Training description} 
They trained their diffusion-based denoising model on a mixed dataset composed of DIV2K and LSDIR, which contained high-resolution images with diverse textures and content. The dataset was augmented with random cropping, horizontal flipping, and other data augmentation techniques to improve model generalization.

The backbone network was selected from either NAFNet, with the feature channel width set to 64. They experimented with different channel sizes and determined that 64 channels provided a good balance between performance and computational efficiency.

They employed theLion optimizer with $\beta_1 = 0.95$ and $\beta_2 = 0.98$ to ensure faster convergence and better stability during training. The learning rate was initialized at $2 \times 10^{-4}$ and was reduced by half after every 200k iterations using a CosineAnnealingLR scheduler to achieve smoother convergence.

The loss function was a Matching Loss designed to minimize the distance between the predicted and true noise residuals. This function integrated L1 and L2 components, weighted dynamically based on the noise variance at different time steps to stabilize the training across different diffusion levels.

They applied mixed precision training with automatic gradient scaling to accelerate training while reducing memory usage. The model was trained for a total of 800k iterations, and each batch contained 16 cropped patches of size $128 \times 128$. Training was conducted using a single NVIDIA RTX 4090 GPU, and the entire process took approximately 36 hours to complete.

To ensure robust noise modeling, a cosine noise schedule was adopted, which progressively adjusted the noise level throughout the training process, allowing the model to better capture high-frequency details during the denoising phase.

\vspace{5pt}

\textbf{Testing description} 
During the training phase, they validated the model using the official validation dataset provided by the NTIRE 2025 competition. The validation set included images with Gaussian noise of varying intensities, and the model was assessed based on both PSNR and SSIM metrics.

Upon completing 800k iterations, the model achieved a peak PSNR of 26.83 dB and an SSIM of 0.79 on the validation dataset, indicating effective noise suppression and structure preservation.

After training was completed, the model was rigorously tested using the official test set to verify its effectiveness in real-world scenarios. They conducted multiple test runs with different noise levels to ensure model robustness across various conditions. The test results confirmed that the model performed consistently well in Gaussian noise removal, maintaining high PSNR and SSIM values across diverse image types.

To further evaluate the performance, they applied both SDE-based and ODE-based sampling methods during inference. ODE sampling provided a faster and more deterministic denoising process, while SDE sampling yielded more diverse results. The final submitted model leveraged ODE sampling to achieve a balance between quality and inference speed.


\subsection{mygo}

U-Net adopts a typical encoder-decoder structure. The encoder is responsible for downsampling the input image, extracting features at different scales to capture the global information and semantic features of the image. The decoder performs upsampling, restoring the feature maps to the original image size and progressively recovering the detailed information of the image. This architecture enables U-Net to achieve rich global semantic information while accurately restoring image details when processing high-definition images, thereby realizing high-precision segmentation.

The U-Net architecture is characterized by its symmetric encoder-decoder structure with skip connections. In the encoder (or contracting path), the network progressively downsamples the input image through multiple convolutional layers interspersed with max-pooling operations. This process allows the model to extract hierarchical features at various scales, capturing both the global context and semantic information of the image.

In the decoder (or expansive path), the network employs transposed convolutions (or upsampling layers) to gradually upscale the feature maps back to the original image resolution. During this process, the decoder receives additional information from the encoder via skip connections, which concatenate corresponding feature maps from the encoder to those in the decoder. This mechanism helps in refining the output by incorporating fine-grained details and spatial information, which are crucial for accurate image restoration or segmentation.

This design ensures that U-Net can effectively handle high-resolution images by leveraging both the broad contextual understanding gained from the encoder and the detailed spatial information preserved through the skip connections. Consequently, this dual capability of capturing global semantics and local details makes U-Net particularly powerful for tasks that require precise image segmentation.
The uniqueness of U-Net lies in its skip connections. These skip connections directly transfer feature maps of the same scale from the encoder to the corresponding layers in the decoder. This mechanism allows the decoder to utilize low-level feature information extracted by the encoder, aiding in the better recovery of image details. When processing high-definition images, these low-level features contain abundant edge, texture, and other detail information, which is crucial for accurate image segmentation.

Compared to Fully Convolutional Networks (FCNs), U-Net stands out because of its use of skip connections. FCN is also a commonly used model for image segmentation, but lacks the skip connections found in U-Net, resulting in poorer performance in recovering detailed image information. When processing high-definition images, FCNs can produce blurry segmentation results with unclear edges. In contrast, U-Net can better preserve the details of the image through its skip connections, thereby improving the accuracy of segmentation.

\begin{figure}[htbp]
    \centering
    \includegraphics[width=0.5\textwidth]{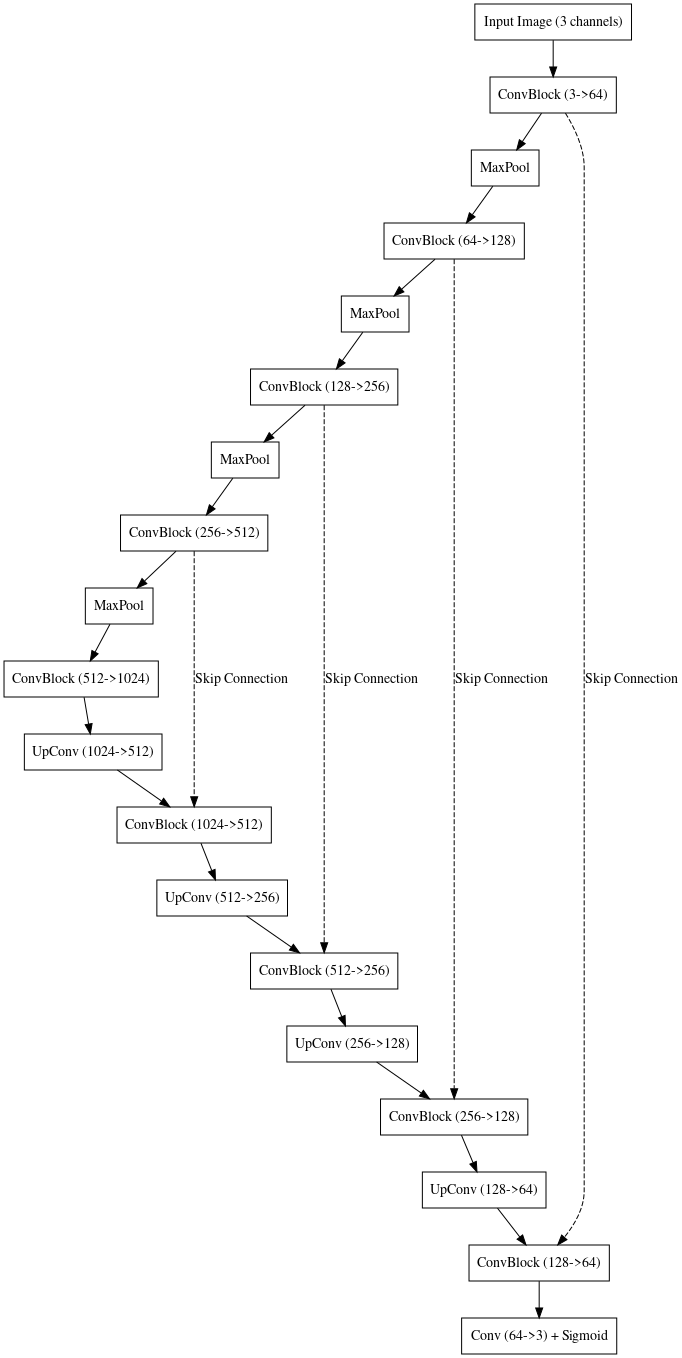}
    \caption{Unet model architecture from Team mygo.}
    \label{fig:example}
\end{figure}

 Our model resizes all images to 512*512 for training, which facilitates the rapid extraction of image features and effectively reduces the usage of video memory (VRAM).
Next, they feed the images into the network model and compute the loss of the output images. In particular, their loss function incorporates both MSE (mean squared error) and SSIM (structured similarity index measure), allowing the model to focus on pixel-level accuracy during training while also emphasizing the structural features of the images. This dual approach improves the overall performance of the model.
They use the Adam optimizer for training, which dynamically adjusts the learning rate during the training process based on the first and second moments of the gradients. This allows it to automatically select the appropriate step sizes for each parameter, leading to more efficient convergence compared to fixed learning rate methods. Additionally, Adam helps reduce the overall memory footprint by maintaining only a few extra parameters per weight, contributing to its efficiency in practical applications. 
In particular, they employ an early stopping mechanism to avoid redundant computations.

It is worth mentioning that they have implemented an early stopping mechanism. This approach helps prevent overfitting by halting the training process when the performance on a validation set stops improving, thus avoiding unnecessary computations and saving computational resources. Early stopping monitors a chosen metric (such as validation loss) and stops training when no improvement is observed over a predefined number of epochs, effectively reducing the risk of overfitting and ensuring efficient use of computational resources.


\section*{Acknowledgments}
This work was partially supported by the Humboldt Foundation, the Ministry of Education and Science of Bulgaria (support for INSAIT, part of the Bulgarian National Roadmap for Research Infrastructure). 
We thank the NTIRE 2025 sponsors: ByteDance, Meituan, Kuaishou, and University of Wurzburg (Computer Vision Lab).

\appendix

\section{Teams and affiliations}
\label{sec:teams}

\subsection*{NTIRE 2025 team}
\noindent\textit{\textbf{Title: }} NTIRE 2025 Image Denoising Challenge\\
\noindent\textit{\textbf{Members: }} \\
Lei Sun$^1$ (\href{mailto:lei.sun@insait.ai}{lei.sun@insait.ai}),\\
Hang Guo$^2$ (\href{mailto:cshguo@gmail.com}{cshguo@gmail.com}),\\
Bin Ren$^{1,3,4}$ (\href{mailto:bin.ren@unitn.it}{bin.ren@unitn.it}),\\
Luc Van Gool$^1$ (\href{mailto:vangool@vision.ee.ethz.ch}{vangool@vision.ee.ethz.ch}),\\
Radu Timofte$^{5}$ (\href{mailto:radu.timofte@uni-wuerzburg.de}{Radu.Timofte@uni-wuerzburg.de})\\
Yawei Li$^6$ (\href{mailto:li.yawei.ai@gmail.com}{li.yawei.ai@gmail.com}),\\
\noindent\textit{\textbf{Affiliations: }}\\
$^1$ INSAIT,Sofia University,"St.Kliment Ohridski”, Bulgaria\\
$^2$ Tsinghua University, China\\
$^3$ University of Pisa, Italy\\
$^4$ University of Trento, Italy\\
$^5$ University of W\"urzburg, Germany\\
$^6$ ETH Z\"urich, Switzerland\\

\subsection*{Samsung MX (Mobile eXperience) Business \& Samsung R\&D Institute China - Beijing (SRC-B)}

\noindent\textit{\textbf{Title: }} Dynamic detail-enhanced image denoising framework\\
\noindent\textit{\textbf{Members: }} \\
Xiangyu Kong$^1$ (\href{mailto:member1@member1.com}{xiangyu.kong@samsung.com}),
Hyunhee Park$^2$, Xiaoxuan Yu$^1$, Suejin Han$^2$, Hakjae Jeon$^2$, Jia Li$^1$, Hyung-Ju Chun$^2$\\
\noindent\textit{\textbf{Affiliations: }} \\ 
$^1$ Samsung R\&D Institute China - Beijing (SRC-B) \\
$^2$ Department of Camera Innovation Group, Samsung Electronics \\
\subsection*{SNUCV}
\noindent\textit{\textbf{Title: }} Deep ensemble for Image denoising\\
\noindent\textit{\textbf{Members: }} \\
Donghun Ryou$^1$ (\href{mailto:dhryou@snu.ac.kr}{dhryou@snu.ac.kr}),
Inju Ha$^1$, Bohyung Han$^1$\\
\noindent\textit{\textbf{Affiliations: }} \\ 
$^1$ Seoul National University \\

\subsection*{BuptMM}
\noindent\textit{\textbf{Title: }} DDU——Image Denoising Unit using transformer and morphology method\\
\noindent\textit{\textbf{Members: }} \\
Jingyu Ma$^1$ (\href{mailto:whalemjy@bupt.edu.cn}{whalemjy@bupt.edu.cn}),
Zhijuan Huang$^2$, Huiyuan Fu$^1$, Hongyuan Yu$^2$, Boqi Zhang$^1$, Jiawei Shi$^1$, Heng Zhang$^2$, Huadong Ma$^1$\\
\noindent\textit{\textbf{Affiliations: }} \\ 
$^1$ Beijing University of Posts and Telecommunications \\
$^2$ Xiaomi Inc., China \\
\subsection*{HMiDenoise}
\noindent\textit{\textbf{Title: }} Hybrid Denosing Method Based on HAT\\
\noindent\textit{\textbf{Members: }} \\
Zhijuan Huang$^1$(\href{mailto:member1@member1.com}{huang\_199109@163.com}),
Jingyu Ma$^2$, Hongyuan Yu$^1$, Heng Zhang$^1$, Huiyuan Fu$^2$, Huadong Ma$^2$\\
\noindent\textit{\textbf{Affiliations: }} \\ 
$^1$ Xiaomi Inc. \\
$^2$ Beijing University of Posts and Telecommunications \\

\subsection*{Pixel Purifiers}
\noindent\textit{\textbf{Title: }} Denoiser using Restormer and Hard Dataset Mining\\
\noindent\textit{\textbf{Members: }} \\
Deepak Kumar Tyagi$^1$ (\href{mailto:deepak.tyagi@samsung.com}{deepak.tyagi@samsung.com}),
Aman Kukretti$^1$, Gajender Sharma$^1$, Sriharsha Koundinya$^1$, Asim Manna$^1$\\
\noindent\textit{\textbf{Affiliations: }} \\ 
$^1$ Samsung R\&D Institute India - Bangalore (SRI-B) \\

\subsection*{Alwaysu}
\noindent\textit{\textbf{Title: }} Bias-Tuning Enables Efficient Image Denoising\\
\noindent\textit{\textbf{Members: }} \\
Jun Cheng $^1$ (\href{mailto:jcheng24@hust.edu.cn}{jcheng24@hust.edu.cn}),
Shan Tan $^1$\\
\noindent\textit{\textbf{Affiliations: }} \\ 
$^1$ Huazhong University of Science and Technology

\subsection*{Tcler\_Denosing}
\noindent\textit{\textbf{Title: }} Tcler Denoising\\
\noindent\textit{\textbf{Members: }} \\
Jun Liu$^{1,2}$ (\href{mailto:jun63.liu@tcl.com}{jun63.liu@tcl.com}),
Jiangwei Hao$^{1,2}$, Jianping Luo$^{1,2}$, Jie Lu$^{1,2}$ \\
\noindent\textit{\textbf{Affiliations: }} \\ 
$^1$ TCL Corporate Research\\
$^2$ TCL Science Park International E City - West Zone, Building D4 \\

\subsection*{cipher\_vision}
\noindent\textit{\textbf{Title: }} Pureformer: Transformer-Based Image Denoising\\
\noindent\textit{\textbf{Members: }} \\
Satya Narayan Tazi$^1$ (\href{mailto:satya.tazi@ecajmer.ac.in}{satya.tazi@ecajmer.ac.in}),
Arnim Gautam$^1$, Aditi Pawar$^1$, Aishwarya Joshi$^2$, Akshay Dudhane$^3$, Praful Hambadre$^4$,  Sachin Chaudhary$^5$, Santosh Kumar Vipparthi$^5$, Subrahmanyam Murala$^6$,\\
\noindent\textit{\textbf{Affiliations: }} \\ 
$^1$ Government Engineering College Ajmer \\
$^2$ Mohamed bin Zayed University of Artificial Intelligence
gence, Abu Dhabi \\
$^3$ University of Petroleum and Energy Studies, Dehradun \\
$^4$ Indian Institute of Technology, Mandi \\
$^5$ Indian Institute of Technology, Ropar \\
$^6$ Trinity College Dublin, Ireland \\

\subsection*{Sky-D}
\noindent\textit{\textbf{Title: }} A Two-Stage Denoising Framework with Generalized Denoising Score Matching Pretraining and Supervised Fine-tuning\\
\noindent\textit{\textbf{Members: }} \\
Jiachen Tu$^1$ (\href{mailto:jtu9@illinois.edu}{jtu9@illinois.edu})

\noindent\textit{\textbf{Affiliations: }} \\ 
$^1$ University of Illinois Urbana-Champaign \\

\subsection*{KLETech-CEVI}
\noindent\textit{\textbf{Title: }} HNNFormer: Hierarchical Noise-Deinterlace Transformer for Image Denoising\\
\noindent\textit{\textbf{Members: }} \\
Nikhil Akalwadi$^{1,3}$ (\href{mailto:nikhil.akalwadi@kletech.ac.in}{nikhil.akalwadi@kletech.ac.in}),
Vijayalaxmi Ashok Aralikatti$^{1,3}$, Dheeraj Damodar Hegde$^{2,3}$, G Gyaneshwar Rao$^{2,3}$, Jatin Kalal$^{2,3}$, Chaitra Desai$^{1,3}$, Ramesh Ashok Tabib$^{2,3}$, Uma Mudenagudi$^{2,3}$\\
\noindent\textit{\textbf{Affiliations: }} \\ 
$^{1}$ School of Computer Science and Engineering, KLE Technological University\\
$^{2}$ School of Electronics and Communication Engineering, KLE Technological University\\
$^{3}$ Center of Excellence in Visual Intelligence (CEVI), KLE Technological University\\

\subsection*{xd\_denoise}
\noindent\textit{\textbf{Title: }} SCUNet for image denoising\\
\noindent\textit{\textbf{Members: }} \\
Zhenyuan Lin$^1$ (\href{mailto:linzhenyuan@stu.xidian.edu.cn}{linzhenyuan@stu.xidian.edu.cn}), 
Yubo Dong$^1$, Weikun Li$^2$, Anqi Li$^1$, Ang Gao$^1$\\
\noindent\textit{\textbf{Affiliations: }} \\ 
$^1$ Xidian University \\
$^2$ Guilin University Of Electronic Technology \\

\subsection*{JNU620}
\noindent\textit{\textbf{Title: }} Image Denoising using NAFNet and RCAN\\
\noindent\textit{\textbf{Members: }} \\
Weijun Yuan$^1$ (\href{mailto:yweijun@stu2022.jnu.edu.cn}{yweijun@stu2022.jnu.edu.cn}),
Zhan Li$^1$, Ruting Deng$^1$, Yihang Chen$^1$, Yifan Deng$^1$, Zhanglu Chen$^1$, Boyang Yao$^1$, Shuling Zheng$^2$, Feng Zhang$^1$, Zhiheng Fu$^1$\\
\noindent\textit{\textbf{Affiliations: }} \\ 
$^1$ Jinan University \\
$^2$ Guangdong University of Foreign Studies \\

\subsection*{PSU\_team}
\noindent\textit{\textbf{Title: }} OptimalDiff: High-Fidelity Image Enhancement Using Schrödinger Bridge Diffusion and Multi-Scale Adversarial Refinement\\

\noindent\textit{\textbf{Members: }} \\
Anas M. Ali$^1$ (\href{mailto:aaboessa@psu.edu.sa}{aaboessa@psu.edu.sa}), 
Bilel Benjdira$^1$, 
Wadii Boulila$^1$\\

\noindent\textit{\textbf{Affiliations: }} \\ 
$^1$ Robotics and Internet-of-Things Laboratory, Prince Sultan University, Riyadh, Saudi Arabia \\

\subsection*{Aurora}
\noindent\textit{\textbf{Title: }} GAN + NAFNet: A Powerful Combination for High-Quality Image Denoising\\
\noindent\textit{\textbf{Members: }} \\
JanSeny (\href{mailto:1225049871@qq.com}{1225049871@qq.com}),
Pei Zhou \\

\subsection*{mpu\_ai}
\noindent\textit{\textbf{Title: }} Enhanced Blind Image Restoration with Channel Attention Transformers and Multi-Scale Attention Prompt Learning\\
\noindent\textit{\textbf{Members: }} \\
Jianhua Hu$^1$ (\href{mailto:p2412994@mpu.edu.mo}{p2412994@mpu.edu.mo}),
K. L. Eddie Law$^1$\\
\noindent\textit{\textbf{Affiliations: }} \\ 
$^1$ Macao Polytechnic University\\

\subsection*{OptDenoiser}

\noindent\textit{\textbf{Title: }} Towards two-stage OptDenoiser framework for image denoising.\\
\noindent\textit{\textbf{Members: }} \\
Jaeho Lee $^1$ (\href{mailto:jaeho.lee@opt-ai.kr}{{jaeho.lee@opt-ai.kr}
}),
M.J. Aashik Rasool$^1$, Abdur Rehman$^1$, SMA Sharif$^1$, Seongwan Kim$^1$\\
\noindent\textit{\textbf{Affiliations: }} \\ 
$^1$ Opt-AI Inc, Marcus Building, Magok, Seoul, South Korea \\

\subsection*{AKDT}
\noindent\textit{\textbf{Title: }} High-resolution Image Denoising via Adaptive Kernel Dilation Transformer\\
\noindent\textit{\textbf{Members: }} \\
Alexandru Brateanu$^1$ (\href{mailto:alexandru.brateanu@student.manchester.ac.uk}{{alexandru\allowbreak.brateanu\allowbreak@student\allowbreak.man\allowbreak chester\allowbreak.ac\allowbreak.uk}
}),
Raul Balmez$^1$, Ciprian Orhei$^2$, Cosmin Ancuti$^2$\\
\noindent\textit{\textbf{Affiliations: }} \\ 
$^1$ University of Manchester - Manchester, United Kingdom \\
$^2$ Polytechnica University Timisoara - Timisoara, Romania\\

\subsection*{X-L}
\noindent\textit{\textbf{Title: }} MixEnsemble\\
\noindent\textit{\textbf{Members: }} \\
Zeyu Xiao$^1$ (\href{mailto:zeyuxiao1997@163.com}{zeyuxiao1997@163.com}),
Zhuoyuan Li$^2$\\
\noindent\textit{\textbf{Affiliations: }} \\ 
$^1$ National University of Singapore \\
$^2$ University of Science and Technology of China \\

\subsection*{Whitehairbin}
\noindent\textit{\textbf{Title: }} Diffusion-based Denoising Model \\

\noindent\textit{\textbf{Members: }} \\
Ziqi Wang$^1$ (\href{mailto:wangziqi-7@outlook.com}{wangziqi-7@outlook.com}), 
Yanyan Wei$^1$, 
Fei Wang$^1$,
Kun Li$^1$, 
Shengeng Tang$^1$, 
Yunkai Zhang$^1$\\

\noindent\textit{\textbf{Affiliations: }} \\ 
$^1$ Hefei University of Technology, China

\subsection*{mygo}
\noindent\textit{\textbf{Title: }} High-resolution Image Denoising via Unet neural network\\
\noindent\textit{\textbf{Members: }} \\
Weirun Zhou$^1$ (\href{1764772710@qq.com}{1764772710@qq.com}),
Haoxuan Lu$^2$ \\

\noindent\textit{\textbf{Affiliations: }} \\ 
$^1$ Xidian University \\
$^2$ China University of Mining and Technology

{\small
\bibliographystyle{ieeenat_fullname}
\bibliography{main}
}

\end{document}